\documentclass{article}
\makeatletter
\g@addto@macro\bfseries{\boldmath}
\makeatother

\usepackage[preprint]{neurips_2026}

\usepackage[utf8]{inputenc} 
\usepackage[T1]{fontenc}    
\usepackage{hyperref}       
\usepackage{url}           
\usepackage{booktabs}      
\usepackage{amsfonts}      
\usepackage{nicefrac}       
\usepackage{microtype}     
\usepackage{xcolor}         
\usepackage{graphicx}

\usepackage{xspace}
\usepackage{amsmath}
\usepackage[capitalize]{cleveref}
\usepackage{multirow}

\usepackage{lineno}
\usepackage{needspace}

\definecolor{denim}{rgb}{0.08, 0.38, 0.74}
\usepackage[table]{xcolor}
\usepackage{enumitem}
\hypersetup{
  colorlinks   = true,
  urlcolor     = magenta, 
  linkcolor    = denim,
  citecolor   =  denim,
}

\definecolor{InferenceRow}{HTML}{EAF3FB}
\definecolor{AgentBRACERow}{HTML}{EAF7EE}

\title{Agent-BRACE: Decoupling Beliefs from Actions in Long-Horizon Tasks via Verbalized State Uncertainty}

\author{\textbf{Joykirat Singh}\textsuperscript{1} \quad
\textbf{Zaid Khan}\textsuperscript{1} \quad
\textbf{Archiki Prasad}\textsuperscript{1} \quad
\textbf{Justin Chih-Yao Chen}\textsuperscript{1} \vspace{3pt} \\ \vspace{7pt}
\textbf{Akshay Nambi}\textsuperscript{3} \quad
\textbf{Hyunji Lee}\textsuperscript{1} \quad
\textbf{Elias Stengel-Eskin}\textsuperscript{2} \quad
\textbf{Mohit Bansal}\textsuperscript{1} \\
\textsuperscript{1}UNC Chapel Hill \quad
\textsuperscript{2}The University of Texas at Austin \quad
\textsuperscript{3}Microsoft Research
}

\usepackage[framemethod=TikZ]{mdframed}
\usepackage{enumitem}
\usepackage{wrapfig}

\newcommand{\shortname}{\texttt{Agent-BRACE}\xspace}
\newcommand{\shortnamenormal}{{Agent-BRACE}\xspace}

\newcommand{\qwentwo}{Qwen2.5-3B-Instruct\xspace}
\newcommand{\qwenthree}{Qwen3-4B-Instruct\xspace}

\definecolor{ExampleBg}{HTML}{ffffff}
\definecolor{ExampleTitle}{HTML}{629677}
\newcounter{example}

\mdfdefinestyle{examplestyle}{
    roundcorner=5pt,
    backgroundcolor=ExampleBg,
    linecolor=ExampleTitle,
    outerlinewidth=0.3pt,
    frametitlebackgroundcolor=ExampleTitle,
    frametitlefont={\bfseries\color{white}},
    skipabove=10pt,
    frametitleaboveskip=3pt,
    frametitlebelowskip=3pt,
    nobreak=true, 
}

\newenvironment{example}[1][]{%
    \refstepcounter{example}%
    \ifstrempty{#1}%
        {\def\exampletitle{Example~\theexample}}%
        {\def\exampletitle{Example~\theexample: #1}}%
    \mdframed[style=examplestyle, frametitle=\exampletitle]%
}{%
    \endmdframed
}

\newenvironment{example*}[1][]{%
    \ifstrempty{#1}%
        {\mdframed[style=examplestyle]}%
        {\mdframed[style=examplestyle, frametitle={#1}]}%
}{%
    \endmdframed
}
\crefname{example}{Example}{Examples}

\definecolor{PromptBg}{HTML}{ffffff}
\definecolor{PromptTitle}{HTML}{2A5FA5}
\newcounter{prompt}
\mdfdefinestyle{promptstyle}{
    roundcorner=5pt,
    backgroundcolor=PromptBg,
    linecolor=PromptTitle,
    outerlinewidth=0.3pt,
    frametitlebackgroundcolor=PromptTitle,
    frametitlefont={\bfseries\color{white}},
    skipabove=10pt,
    frametitleaboveskip=3pt,
    frametitlebelowskip=3pt,
    nobreak=true,
}
\newenvironment{prompt}[1][]{%
    \refstepcounter{prompt}%
    \ifstrempty{#1}%
        {\def\prompttitle{Prompt~\theprompt}}%
        {\def\prompttitle{Prompt~\theprompt: #1}}%
    \mdframed[style=promptstyle, frametitle=\prompttitle]%
}{%
    \endmdframed
}
\newenvironment{prompt*}[1][]{%
    \ifstrempty{#1}%
        {\mdframed[style=promptstyle]}%
        {\mdframed[style=promptstyle, frametitle={#1}]}%
}{%
    \endmdframed
}
\crefname{prompt}{Prompt}{Prompts}

\begin{document}

\maketitle

\begin{abstract}

Large language models~(LLMs) are increasingly deployed on long-horizon tasks in partially observable environments, where they must act while inferring and tracking a complex environment state over many steps. This leads to two challenges: \textit{partial observability} requires maintaining uncertainty over unobserved world attributes, and \textit{long interaction history} causes context to grow without bound, diluting task-relevant information. A principled solution to both challenges is a \textit{belief state}: a posterior distribution over environment states given past observations and actions, which compactly encodes history for decision making regardless of episode length. In LLM agents, however, the open-ended nature of text makes it unclear how to represent such a distribution.
Therefore, we introduce \shortname: \textbf{Agent} \textbf{B}elief state \textbf{R}epresentation via \textbf{A}bstraction and \textbf{C}onfidence \textbf{E}stimation, a method that decouples an LLM agent into a \textit{belief state model} 
and a \textit{policy model}, jointly optimized via reinforcement learning. The belief state model produces a structured approximation of the belief distribution: a set of atomic natural language claims about the environment, each annotated with an ordinal verbalized certainty label ranging from certain to unknown. The policy model conditions on this compact, structured approximate belief rather than the full history, learning to select actions under explicit uncertainty. 
Across long-horizon, partially observable embodied language environments, \shortname achieves an average absolute improvement of +14.5\% (\qwentwo) and +5.3\% (\qwenthree), outperforming strong RL baselines while maintaining a near-constant context window independent of episode length. Further analysis shows that the learned belief becomes increasingly calibrated over the course of an episode as evidence accumulates.\footnote{Codebase: \href{https://github.com/joykirat18/Agent-BRACE}{https://github.com/joykirat18/Agent-BRACE}}
\end{abstract}

\section{Introduction}

\begin{figure}
    \centering
    \includegraphics[width=\textwidth]{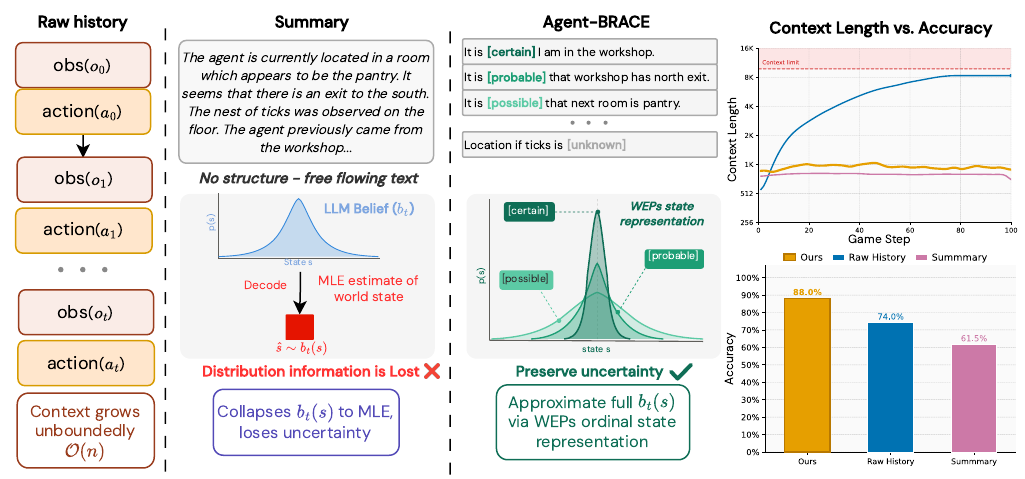}
    \caption{Three approaches to context management in long-horizon POMDP agents. \textbf{Raw history} (left), grows unbounded as $\mathcal{O}(n)$. \textbf{Summary-based belief} (middle) compresses history into a summary but collapses the belief distribution to a single MLE point estimate \(\hat{s} \sim b_t(s)
\), discarding uncertainty. \textbf{\shortnamenormal} (right) represents the belief as WEP-annotated statements (\textit{confirmed}, \textit{probable}, \textit{possible}, etc.), approximating the full distribution $b_t(s)$, with near-constant context window. \shortname (\qwenthree) outperforms both baselines in accuracy while maintaining constant context length (right panel).}
    \label{fig:intro}
    \vspace{-0.5em}
\end{figure}

Large language models (LLMs) are increasingly being deployed as agents in long-horizon, partially observable tasks like software engineering~\citep{yang2024sweagentagentcomputerinterfacesenable, jimenez2024swebenchlanguagemodelsresolve}, web navigation~\citep{zhou2023webarena, deng2023mind2web, he2024webvoyager}, or research~\citep{lu2024ai, novikov2025alphaevolve}.
These models must act while inferring complex world state from incomplete observations over many steps -- a setting that is traditionally modeled as a partially observable Markov Decision Process (POMDP)~\citep{aastrom1965optimal}.
In this framing, an optimal policy needs to only condition on the \textit{belief state}, a posterior distribution over possible environment states given the history of past/current observations and past actions.
The belief state admits two complementary interpretations: \textit{(1)} the distribution represents uncertainty over the state the agent is in, accounting for unobserved variables; \textit{(2)} it serves as a sufficient statistic for the prior interaction history $\mathcal{H}_t$, allowing the agent to track observations over time.
Current LLM agents differ from traditional POMDP approaches in that they generally represent both actions and observations in text.
This enables interaction with open-ended, unstructured environments that lack predefined action or observation schema, but complicates encoding an explicit belief state and introduces its own challenges. 
First, without a sufficient statistic of history, LLM-based policies must be conditioned on the raw interaction trajectory, leading to inefficient representation~(\cref{fig:intro};~\textit{Raw history}), with the context length growing linearly in the episode length, thus increasing computational cost and diluting task-relevant signals with spurious details~\citep{liu2024lost, chung2025evaluating}~(\cref{fig:intro}; \textit{Context Length vs  Accuracy}). 
Second, while POMDP approaches for large or continuous state spaces are well studied (e.g., particle filters, predictive state representations) \citep{silver2010monte, hafner2020mastering, gregor2019shaping}, the open-ended nature of text poses its own challenges: it is unclear how to encode a distribution in text over a compositional state space. 
Indeed, past work either relies on the LLM's internal representation as a belief proxy~\citep{kamel2025emergent} -- which lacks interpretability and limits external verification -- or externalizes belief into a free-form natural language summary \citep{zhou2025mem1, yu2025memagent}, which is more interpretable but collapses the belief distribution $b_t(s)$ into a single point estimate.

\begin{figure}
    \centering
    \includegraphics[width=\linewidth]{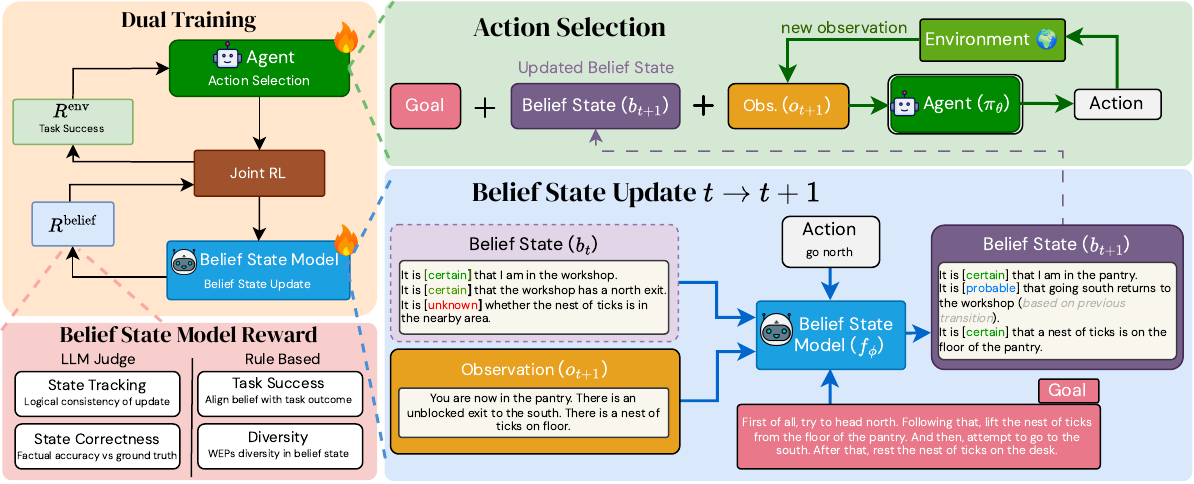}
\label{fig:brace_overview}
    \caption{
    Overview of \shortname. The agent is decomposed into a \textit{belief state model} $f_\phi$ and a \textit{policy model} $\pi_\theta$, jointly optimized via PPO (\textit{Dual Training}). At each step $t$, $f_\phi$ consumes the goal $G$, previous belief $b_{t}$, and new observation $o_{t+1}$ to produce an updated belief $b_{t+1}$ with WEPs-based certainty labels (\textit{Belief State Update}). The policy $\pi_\theta$ then selects an action $a_t$ conditioned on $(G, b_{t+1}, o_{t+1})$ rather than the full history $\mathcal{H}_t$ (\textit{Action Selection}). The belief model is trained with a composite reward $R^{\text{belief}}$, while the policy model is trained with a binary environment reward $R^{\text{env}}$. 
    }
    \label{fig:method}
\vspace{-0.5em}
\end{figure}

To tackle these challenges and preserve uncertainty in belief states for LLM agents, we introduce \shortname: \textbf{Agent} \textbf{B}elief state \textbf{R}epresentation via \textbf{A}bstraction and \textbf{C}onfidence \textbf{E}stimation, a training method that represents an agent's belief as text while simultaneously encoding uncertainty via verbalized probability estimates. 
Following the POMDP formalization, \shortname decouples an LLM agent into two modules: a \textit{belief state model} and a \textit{policy model}, training them jointly using reinforcement learning (RL).
As shown in ~\cref{fig:method} (\textit{belief state update}), at each step $t$ the belief state model takes as input the goal ($G$), the previous belief ($b_t$), and the new observation ($o_{t+1}$), and produces an updated approximate belief ($b_{t+1}$) represented as a set of atomic natural language claims. 
Crucially, each claim is annotated with a certainty label drawn from the Words of Estimative Probability (WEP) scale~\citep{10.1162/opmi_a_00066, tang2026evaluation, sileo2023probing}; an ordered Likert style vocabulary (\textit{confirmed} $\succ$ \textit{almost certain} $\succ$ \textit{probable} $\succ$ \textit{possible} $\succ$ \textit{unlikely} $\succ$\textit{doubtful} $\succ$ \textit{unknown}) that is grounded in how humans express uncertainty in natural language.
Prior work has shown LLMs can meaningfully produce and differentiate between such verbalized uncertainty expressions~\citep{lin2022teaching, tian2023just, stengel2024lacie}. 
This yields a belief approximation that
captures uncertainty and uses a discrete scale that LLMs can reliably produce and update. 
Since the belief state $b_t$ is a sufficient approximation of the full history,  the policy model can select an action conditioned on $(G, b_t, o_t)$ rather than on the history $\mathcal{H}_t$ -- replacing an ever-growing trajectory with a compact, bounded representation. 
In \shortname, the belief state model and the policy model are jointly trained via PPO~\citep{schulman2017proximalpolicyoptimizationalgorithms}. 
The policy model is optimized to maximize the binary environment reward (+1 for success, 0 for failure), providing the primary signal for action selection. The belief state model is optimized via a combination of complementary rewards, each targeting a different failure mode in belief quality: state tracking ensures logical consistency~\citep{zou2026reducingbeliefdeviationreinforcement, yuan2026rpms}, state correctness ensures factual grounding~\citep{zhao2026gracereinforcementlearninggrounded}, diversity prevents uncertainty collapse~\citep{leng2024taming}, discounted success aligns belief quality with task outcomes, and format ensures structural consistency. Ablations in Section~\ref{sec:ablation} confirm the importance of each reward. 

We train and evaluate \shortname on various long-horizon, partially observable embodied language tasks. Specifically, \shortname is trained on Quest, a task from the \textit{TextWorld}~\citep{cote2018textworld} environment, using \qwentwo~\citep{qwen2.5} and \qwenthree~\citep{qwen3technicalreport} as base models and evaluate on three \textit{TextWorld} environments: Quest, Treasure, and Cooking.
\shortname outperforms all baselines, including ReAct~\citep{yao2022react}, Direct-Action (RL trained), ReAct (RL trained), MEM1~\citep{zhou2025mem1}, and PABU~\citep{jiang2026pabu}, achieving average accuracies of 72.8\% and 79.3\% on \qwentwo and \qwenthree, respectively -- an average absolute improvement\footnote{All improvements reported in this paper are absolute unless otherwise stated.} of +14.5\% over the strongest RL-trained baseline (Direct-Action (RL)) on \qwentwo and +5.3\% on \qwenthree. Crucially, \shortname maintains a near constant context window while achieving the best performance.
\shortname also demonstrates strong generalization, achieving consistently high performance on Treasure and Cooking tasks despite being trained only on Quest. Moreover, we show that \shortname can be extended to other tasks, with +2.85\% improvement over the strongest RL-trained baseline on ALFWorld~\citep{shridhar2020alfworld}.
Our ablations confirm that each component contributes meaningfully: joint training, belief-state rewards, and an expressive WEP label set each play a critical role -- removing any one leads to meaningful degradation.
Further analysis shows that the belief becomes better calibrated over the course of an episode, with Brier score~\citep{glenn1950verification} decreasing from 0.40 to 0.28 and the fraction of \textit{confirmed} claims growing from 21\% to 52\% as evidence accumulates.

\section{Methodology: \shortnamenormal}

In this section, we introduce our method in detail~(\cref{fig:method}). \shortname addresses two core challenges in long-horizon agentic tasks: linear growth of past history and the absence of belief representation under partial observability. To tackle this, \shortname jointly trains a \textit{belief state model} and a \textit{policy model} via PPO, where the belief state model maintains a structured uncertainty-aware belief that serves as a sufficient approximation of the history for downstream action selection.

\subsection{Environment and Agentic Task}
\label{method: env_agentic_task}
We focus on partially-observable environments modeled as POMDPs defined by the tuple $\mathcal{M} = (S, T, A, \Omega, O, R, \gamma)$, where $S$ is the set of latent environment states, $T: S \times A \to \Delta(S)$ is the state-transition distribution, 
$A$ is the natural language action space,  $\Omega$ is the observation space, $O: S \times A \to \Delta(\Omega)$ is the observation distribution, $R: S \times A \to \mathbb{R}$ is the reward function, and $\gamma \in (0, 1)$ is the discount factor. Since the current observation $o_t$ is not a sufficient statistic for the environment state $s_t$ \citep{kaelbling1998planning}, an optimal policy $\pi$ must condition on the full history $\mathcal{H}_t = \{G, o_0, a_0, \dots, o_t\}$, or an equivalent belief state $b_t = P(s_t | \mathcal{H}_t)$, to maximize expected cumulative rewards~\citep{aastrom1965optimal}. 
We consider agentic tasks where an LLM pursues a goal $G$ by interacting with the environment $\epsilon$, until the objective is achieved or a step budget is reached.

\subsection{Decoupled Architecture: Belief State Model and Policy Model}
\label{method: decoupled_architecture}
As shown in \cref{fig:method} (\textit{Dual Training}), our approach parameterizes an agent with two jointly optimized components: a \textit{belief state model} for state estimation and a \textit{policy model} for action selection: 

\textbf{Belief State Model ($f_\phi$):} This model is a learnable belief-update function, constructing and maintaining an approximate belief representation from raw environment observations. Each belief state is represented as a set of statements, where each statement is annotated with a WEP-based uncertainty label. As shown in ~\cref{fig:method} (\textit{Belief State Update}), $f_\phi$ consumes the goal $\textit{G}$, the current belief state $b_{t}$, and the new observation $o_{t+1}$ to produce an updated belief state \( b_{t+1} = f_\phi(\textit{G}, b_{t}, o_{t+1})\).

\textbf{Policy Model ($\pi_\theta$):} This targets the long-horizon history challenge by conditioning action selection on the compact belief $b_{t+1}$ generated by the belief state model rather than the full history $\mathcal{H}_t$. The next action becomes \(\pi_\theta(\textit{G}, b_{t+1}, o_{t+1})\).

\subsection{Belief State Representation}
\label{method: belief_state}

As shown in \cref{fig:intro} (\textit{Summary}), representing belief states as natural language summaries collapses the distribution over environment state $s_t$ into a single point estimate, discarding the \textit{uncertainty} that belief state policies rely on in partially observable settings. Such summaries are unstructured, producing free-flowing prose with no separation between distinct facts, making it difficult for the policy model to locate and extract task-relevant signals. This problem compounds over time as summaries grow to accommodate new observations~\citep{kang2025acon}, increasing both the length and density of interleaved facts the policy must parse. 
Instead, we represent belief state $b_t$ as \textit{a set of verbalized belief statements}: atomic natural language claims about individual aspects of the environment, each annotated with an explicit uncertainty label~(example shown in \cref{fig:method}) drawn from the Words of Estimative Probability (WEP) scale~\citep{Kent1964-KENWOE} 
-- an ordered Likert-style vocabulary grounded in how human naturally express uncertainty: \textit{confirmed} $\succ$ \textit{almost certain} $\succ$ \textit{probable} $\succ$ \textit{possible} $\succ$ \textit{unlikely} $\succ$\textit{doubtful} $\succ$ \textit{unknown}. This yields a structured approximation of the belief distribution $b_t(s)$ that is both interpretable and reliably produced by LLMs~\citep{lin2022teaching, tian2023just, xiong2023can}.

To define the belief space, we specify set of structured slots that every belief state must populate (e.g, agent location, inventory, etc.); further details in Appendix~\ref{app:belief_structure}.
While specific values and their instantiations are learned from task experience rather than hand-specified, automatically discovering belief dimensions in fully open-ended environments remains future work. We ablate this structured belief space in Appendix~\ref{app:belief_without_task}, finding only a minor drop in performance without it, suggesting that the joint training provides a strong foundation for belief learning, with structured belief states delivering additional gains.

\subsection{Joint Training Procedure}
\label{method: joint_training}
As shown in ~\cref{fig:method}, we train the belief state model $f_\phi$ and policy model $\pi_\theta$ jointly via Proximal Policy Optimization (PPO)~\citep{schulman2017proximalpolicyoptimizationalgorithms}. 
Prior to PPO, we perform supervised finetuning~(SFT) on the belief state model using GPT 5.4 mini~\citep{gpt54} trajectories to enforce a structured belief state representation (SFT detail in Appendix~\ref{app: SFT}; SFT stage ablation in Appendix~\ref{app:belief_without_task}). 
The policy model $\pi_\theta$ is trained to maximize the binary environment reward $R^{\text{env}}$.
Simultaneously, the belief state model is optimized to maximize the composite reward $R^{\text{belief}}$. The clipped objective is given by:
\begin{equation}
    \mathcal{L}_{\text{policy/belief}}(\theta) = -\hat{\mathbb{E}}_t \left[ \min\left( r_t(\theta)\hat{A}_t, \text{clip}(r_t(\theta), 1-\epsilon, 1+\epsilon)\hat{A}_t \right) \right]
\end{equation}
where $r_t(\theta)$ is the probability ratio. For the policy model, $\hat{A}_t$ is estimated via Generalized Advantage Estimation (GAE)~\citep{schulman2015high} using a learned critic $V(s_t)$. To reduce the computational cost, we approximate the advantage: $\hat{A}_t^{\text{belief}} = \frac{R_t^{\text{belief}} - \mu_r}{\sigma_r}$, where $\mu_r$  and $\sigma_r$ are the mean and standard deviation of rewards computed over rollouts from the same task. Both modules are trained via the same objective, but differ in advantage estimation: the policy model uses GAE with a learned critic $V(s_t)$, while the belief state model uses GRPO style group-normalized returns~\citep{shao2024deepseekmath}. 

\subsection{Reward Design}
\label{method:reward_design}
\shortname is trained with two distinct reward signals: $R^{\text{belief}}$ for the belief state model and $R^{\text{env}}$ for the policy model, optimized jointly so that the belief representations are shaped by the policy's decision-making needs. Both rewards are summarized in ~\cref{tab:reward_design}. The final belief state model reward \(R^{\text{belief}}_t = r^{\text{format}}_t \times \frac{1}{4}\bigl(r^{\text{st}}_t + r^{\text{sc}}_t + r^{\text{div}}_t + r^{\text{success}}_t\bigr)\). Full detail in Appendix~\ref{app:reward_design}.

\begin{table}[h]
\centering
\small
\caption{Reward components for the belief state model and policy model.}
\begin{tabular}{l|l|p{8.5cm}}
\toprule
\textbf{Reward} & \textbf{Symbol} & \textbf{Purpose} \\ \midrule
\multicolumn{3}{c}{Belief State Model ($R^{\text{belief}}$)} \\ \midrule
State Tracking    & $r_t^{\text{st}}$       & Measures the logical consistency of the belief update $b_{t-1} \to b_t$ given observation $o_t$. \\
State Correctness & $r_t^{\text{sc}}$       & Ensures that the belief state claim, along with its uncertainty score, remains grounded to the environment states. \\
Diversity         & $r_t^{\text{div}}$      & Encourages use of the full WEP vocabulary via entropy $H(b_t)$ of the label histogram \\
Format            & $r_t^{\text{format}}$   & Enforces structured output; acts as a multiplicative gate zeroing all other rewards for invalid outputs \\
Task Success & $r_t^{\text{success}}$ & Propagates task outcome to the belief module via $\gamma^t \times \mathbf{1}[\text{success}]$ \\
\midrule
\multicolumn{3}{c}{Policy Model ($R^{\text{env}}$)} \\
\midrule
Task Success      & $R^{\text{env}}$        & Binary reward from the environment ($+1$ for success, $0$ for failure); primary reinforcement signal for action selection \\
\bottomrule
\end{tabular}
\label{tab:reward_design}
\end{table}

\section{Experimental Setup and Results}

\subsection{Setup}
\label{sec:experimental_setup}
\textbf{Models.} To evaluate \shortname, we adopt instruction tuned models, \qwentwo~\citep{qwen2.5} and \qwenthree~\citep{qwen3technicalreport} as our base models. Both the belief state and policy models are initialized from the same base model. 
Additionally, we use Qwen3-30B-A3B-Instruct~\citep{qwen3technicalreport} as the judge to evaluate state tracking and correctness reward.

\textbf{Datasets.} We train and evaluate on \textit{TextWorld}~\citep{cote2018textworld}, which provides the flexibility to generate multiple types of text-based games.
We construct three different tasks using \textit{TextWorld} environment:
(i) \textbf{Quests:} A text adventure game environment where agents navigate rooms, manipulate objects, and solve quests via natural language; (ii) \textbf{Cooking:} This task 
takes place in a typical house and consists in finding the right food item and cooking it; (iii) \textbf{Treasure:} The agent spawns in a randomly generated maze and must find a specific object which is mentioned in the objective displayed when the game starts. \shortname and other training baselines were trained only on Quest and evaluated on all three datasets. Full details are provided in Appendix~\ref{app:dataset}. 

\textbf{Baselines.} We compare with several strong baselines \textbf{(1) Base Model:} off the shelf instruction tuned model; \textbf{(2) ReAct}~\citep{yao2022react}: Interleaved reasoning and action selection; \textbf{(3) Direct-Action (RL):} PPO trained model that directly outputs actions, using the same final environment reward as \shortname; \textbf{(4) ReAct (RL):} PPO trained model that additionally outputs its thinking inside \textit{<think>} ... \textit{</think>} tokens before taking an action; \textbf{(5) MEM1}~\citep{zhou2025mem1}: RL framework that maintains a compact shared state for memory consolidation and reasoning -- integrating prior memory with new observations while strategically discarding irrelevant or redundant information; \textbf{(6) PABU}~\citep{jiang2026pabu}: Belief-state framework that compactly represents an agent’s state by explicitly modeling task progress and selectively retaining past actions and observation.

\textbf{Implementation Details.} The maximum number of turns during training is set to be 15, and during inference, to test the long-horizon capability of the method, we set the maximum number of turns to be 100. Additional hyperparameter details are available in Appendix~\ref{app:hyperparameters}.

\subsection{Main Results}

\begin{table}
\centering
\small
\caption{
Performance comparison of \shortname against baselines across three TextWorld
environments (Quest, Treasure, Cooking) on \qwentwo and \qwenthree. Acc.\ denotes task
success rate (\%, higher is better); Steps denotes average number of steps taken (lower
is better). \colorbox{InferenceRow}{Blue rows} are inference-only (no training); \shortname and other baselines are
trained with Quest only, while Treasure and Cooking are out-of-domain (OOD) tasks.
}
\begin{tabular}{lllllllll}
\toprule
\multicolumn{1}{l|}{Method} & \multicolumn{2}{c|}{Quest} & \multicolumn{2}{c|}{Treasure} & \multicolumn{2}{c|}{Cooking} & \multicolumn{2}{c}{Average} \\ \midrule
\multicolumn{1}{l|}{} & Acc.$\uparrow$ & \multicolumn{1}{l|}{Steps$\downarrow$} & Acc.$\uparrow$ & \multicolumn{1}{l|}{Steps$\downarrow$} & Acc.$\uparrow$ & \multicolumn{1}{l|}{Steps$\downarrow$} & Acc.$\uparrow$ & Steps$\downarrow$ \\ \midrule
\multicolumn{9}{c}{\textbf{\qwentwo}} \\ \midrule
\rowcolor{InferenceRow}\multicolumn{1}{l|}{Base Model}         & 4.0  & \multicolumn{1}{l|}{96.1} & 7.5  & \multicolumn{1}{l|}{93.2} & 2.5  & \multicolumn{1}{l|}{98.1} & 4.7  & 95.8 \\
\rowcolor{InferenceRow}\multicolumn{1}{l|}{ReAct}              & 23.0 & \multicolumn{1}{l|}{37.6} & 37.0 & \multicolumn{1}{l|}{33.6} & 27.5 & \multicolumn{1}{l|}{38.4} & 29.2 & 36.5 \\
\multicolumn{1}{l|}{Direct-Action (RL)}                        & 56.0 & \multicolumn{1}{l|}{35.8} & 67.5 & \multicolumn{1}{l|}{32.6} & 51.5 & \multicolumn{1}{l|}{46.1} & 58.3 & 38.2 \\
\multicolumn{1}{l|}{ReAct (RL)}                                & 46.5 & \multicolumn{1}{l|}{34.2} & 55.0 & \multicolumn{1}{l|}{32.7} & 34.5 & \multicolumn{1}{l|}{44.4} & 45.3 & 37.1 \\
\multicolumn{1}{l|}{MEM1}                                      & 29.5 & \multicolumn{1}{l|}{62.9} & 30.0 & \multicolumn{1}{l|}{47.7} & 52.5 & \multicolumn{1}{l|}{48.0} & 37.3 & 52.9 \\
\multicolumn{1}{l|}{PABU}                                      & 73.0 & \multicolumn{1}{l|}{37.0} & 72.5 & \multicolumn{1}{l|}{34.4} & 33.0 & \multicolumn{1}{l|}{73.1} & 59.5 & 48.2 \\
\multicolumn{1}{l|}{\shortname}        & \textbf{78.5} & \multicolumn{1}{l|}{37.3} & \textbf{81.5} & \multicolumn{1}{l|}{32.1} & \textbf{58.5} & \multicolumn{1}{l|}{60.3} & \textbf{72.8} & 43.3 \\ \midrule
\multicolumn{9}{c}{\textbf{\qwenthree}} \\ \midrule
\rowcolor{InferenceRow}\multicolumn{1}{l|}{Base Model}         & 61.5 & \multicolumn{1}{l|}{32.3} & 65.0 & \multicolumn{1}{l|}{30.3} & 69.5 & \multicolumn{1}{l|}{34.1} & 65.3 & 32.2 \\
\rowcolor{InferenceRow}\multicolumn{1}{l|}{ReAct}              & 60.5 & \multicolumn{1}{l|}{12.6} & 69.5 & \multicolumn{1}{l|}{10.3} & 13.5 & \multicolumn{1}{l|}{24.4} & 47.8 & 15.8 \\
\multicolumn{1}{l|}{Direct-Action (RL)}                        & 74.0 & \multicolumn{1}{l|}{29.6} & 72.5 & \multicolumn{1}{l|}{28.0} & \textbf{75.5} & \multicolumn{1}{l|}{31.9} & 74.0 & 29.8 \\
\multicolumn{1}{l|}{ReAct (RL)}                                & 75.5 & \multicolumn{1}{l|}{18.2} & 74.0 & \multicolumn{1}{l|}{16.5} & 13.0 & \multicolumn{1}{l|}{40.6} & 54.2 & 25.0 \\
\multicolumn{1}{l|}{MEM1}                                      & 61.5 & \multicolumn{1}{l|}{50.2} & 63.5 & \multicolumn{1}{l|}{31.4} & 10.0 & \multicolumn{1}{l|}{10.0} & 45.0 & 30.5 \\
\multicolumn{1}{l|}{PABU}                                      & 82.2 & \multicolumn{1}{l|}{29.1} & 73.5 & \multicolumn{1}{l|}{37.2} & 32.5 & \multicolumn{1}{l|}{75.6} & 62.7 & 47.3 \\
\multicolumn{1}{l|}{\shortname}        & \textbf{88.0} & \multicolumn{1}{l|}{30.5} & \textbf{81.0} & \multicolumn{1}{l|}{30.0} & 69.0 & \multicolumn{1}{l|}{44.6} & \textbf{79.3} & 35.0 \\ \toprule
\end{tabular}
\label{tab:main_result}
\end{table}

\textbf{\shortnamenormal outperforms other baselines.}
\cref{tab:main_result} presents the main results across three \textit{TextWorld} environments for both \qwentwo and \qwenthree. 
Overall, \shortname achieves the highest average accuracy across all baselines with 72.8\% on \qwentwo and 79.3\% on \qwenthree, an absolute improvement of +14.5\% and +5.3\% over the strongest RL-trained baseline, Direct-Action (RL), respectively. 
On \qwentwo, \shortname outperforms ReAct (RL) by +27.5\%, demonstrating that interleaved chain-of-thought reasoning alone is insufficient under partial observability. Against MEM1, \shortname improves by +35.5\% on \qwentwo and +34.3\% on \qwenthree, confirming that summary-based compression discards task-critical signals. \shortname on average also outperforms PABU by +13.3\% on \qwentwo and +16.6\% on \qwenthree. The improvements are consistent across both models and suggest a clear pattern: baselines that treat history as a sufficient statistic, whether through raw context (Direct Action, ReAct), summarization (MEM1), or progress-aware compression (PABU), cannot maintain an uncertainty approximation over world state. \shortname's improvement stems from explicit representation of an approximate belief state via WEP annotations and jointly optimizing the belief state model with the policy, so that the agent learns to act under uncertainty rather than from a single point estimate of the world. 

\begin{figure}
    \centering
    \includegraphics[width=0.8\textwidth]{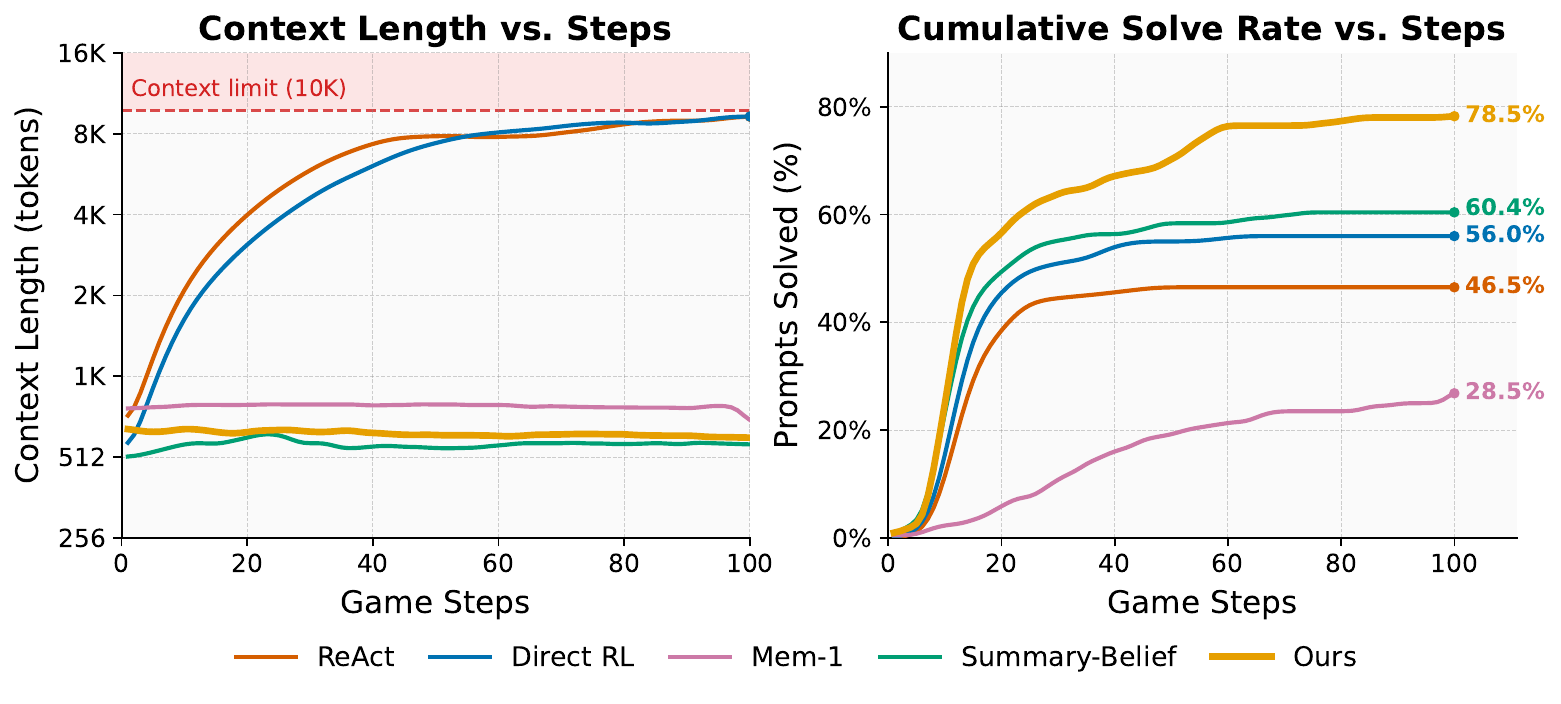}
    \caption{\shortname maintains a near constant context window while achieving the highest solve rate (78.5\%). Comparison of context length growth (left) and cumulative solve rate (right) across methods with maximum 100 game steps on Quest using \qwentwo. 
    }
    \label{fig:analysis}
\vspace{-1.5em}
\end{figure}

\textbf{\shortnamenormal generalizes to held-out TextWorld Environments.}
\shortname is exclusively trained on Quest, yet achieves strong and consistent performance across all three tasks (~\cref{tab:main_result}). It transfers most effectively to Treasure (81.5\% on \qwentwo and 81.0\% on \qwenthree), which shares Quest's navigation structure. On cooking, which requires sequential sub-goal completion and is structured differently from Quest, \shortname still outperforms most baselines (+7.0\% over Direct-Action and +6.0\% over MEM1 on \qwentwo). MEM1, ReAct (RL), and PABU all collapse on Cooking under \qwenthree (10.0\%, 13.0\%, and 32.5\%), confirming that history-based and progress-aware representations are brittle under the non-monotonic sub-goal structure of Cooking. \shortname, maintaining per-claim uncertainty without assuming a linear progress signal, remains robust across both settings without per-task engineering.
 
\textbf{\shortnamenormal maintains a bounded context window and enhanced solve rate.} ~\cref{fig:analysis} plots context length (left) and cumulative solve rate (right) across methods over 100 steps. ReAct and Direct-Action (RL) grow linearly, eventually exceeding the context limit, while \shortname maintains a near constant context window. Crucially, despite operating within a similar context budget as MEM1 and Summary-Belief (ablation run where belief state model is replaced with LLM summarizer), \shortname achieves a substantially higher cumulative solve rate of 78.5\% vs  28.5\% and 60.4\% respectively. The gap isolates the contribution of approximating the belief distribution via WEP labels from context bounding alone.

\begin{wraptable}{r}{0.42\textwidth}
\centering
\small
\caption{Performance comparison of \shortname (\qwenthree) against baselines on ALFWorld environment. \colorbox{InferenceRow}{Blue rows} are inference-only. }
\begin{tabular}{lll}
\toprule
\multicolumn{1}{l|}{Method} & Acc.$\uparrow$ & \multicolumn{1}{l}{Steps$\downarrow$} \\ \midrule
\rowcolor{InferenceRow}\multicolumn{1}{l|}{Base Model}         &  24.3 & \multicolumn{1}{l}{20.1} \\
\rowcolor{InferenceRow}\multicolumn{1}{l|}{ReAct}              & 16.4 & \multicolumn{1}{l}{9.6} \\
\multicolumn{1}{l|}{Direct-Action (RL)}                        & 27.9 & \multicolumn{1}{l}{19.1}  \\
\multicolumn{1}{l|}{ReAct (RL)}                                & 22.1 & \multicolumn{1}{l}{10.6} \\
\multicolumn{1}{l|}{MEM1}                                &  25.7 & \multicolumn{1}{l}{42.9} \\
\multicolumn{1}{l|}{\shortname}        & \textbf{30.7} & \multicolumn{1}{l}{39.1} \\ \toprule
\end{tabular}
\label{tab:alfworld_result}
\end{wraptable}

\textbf{\shortnamenormal performance on ALFWorld.} While Quest, Cooking, and Treasure share common navigation and inventory structure, we wanted to test \shortname generalization to tasks with qualitatively different actions. ALFWorld~\citep{shridhar2020alfworld} is built on top of the ALFRED~\citep{ALFRED20} household dataset that requires the agent to execute multi-step object manipulation (e.g., pick, clean, heat, cool, place) with different observation and action structures from the TextWorld environment. ~\cref {tab:alfworld_result} shows the performance of \shortname (\qwenthree) on ALFWorld, averaged over three evaluation runs. \shortname achieves the highest accuracy of 30.71\%, outperforming the strongest RL-trained baseline Direct-Action (RL) by +2.85\% and the Base Model by +6.42\%. It also outperforms memory based baseline MEM1 by 5\%. These results suggest that \shortname's structured belief representation generalizes to other tasks.

\section{Ablation and Analysis}
\label{sec:ablation}
To understand the importance of each component of \shortname, we do an ablation study, specifically \textbf{(1) \shortname (Limited WEP):} Training the belief state model to only capture two levels of uncertainty -- \textit{confirmed} and \textit{unknown}; \textbf{(2) \shortname (Summary-Belief)}: Instead of training belief state to capture both abstraction of past history and uncertainty, it only summarizes past history $\mathcal{H}_k$; \textbf{(3) \shortname (Frozen Belief model):} The belief state model remains static and only the policy model is trained; \textbf{(4) \shortname (No State Reward):} The belief model is only trained on discounted success reward and other state relevant rewards are removed. 

\begin{table}
\centering
\small
\caption{Ablation analysis of \shortname using \qwentwo and \qwenthree. Each component is a variation of \shortname.
}
\begin{tabular}{lllllllll}
\toprule
\multicolumn{1}{l|}{Method}              & \multicolumn{2}{c|}{Quest}            & \multicolumn{2}{c|}{Treasure}         & \multicolumn{2}{c|}{Cooking}          & \multicolumn{2}{c}{Average} \\ \midrule
\multicolumn{1}{l|}{}                    & Acc.$\uparrow$ & \multicolumn{1}{l|}{Steps$\downarrow$} & Acc.$\uparrow$ & \multicolumn{1}{l|}{Steps$\downarrow$} & Acc.$\uparrow$ & \multicolumn{1}{l|}{Steps$\downarrow$} & Acc.$\uparrow$       & Steps$\downarrow$      \\ \midrule
\multicolumn{9}{c}{\textbf{Qwen-2.5-3B-Instruct}} \\ \midrule

\rowcolor{AgentBRACERow}\multicolumn{1}{l|}{\shortname}          &   \textbf{78.5}& \multicolumn{1}{l|}{37.3}      &      \textbf{81.5}& \multicolumn{1}{l|}{32.1}      &       58.5& \multicolumn{1}{l|}{60.3}      &        \textbf{72.8}&     43.3\\
\multicolumn{1}{l|}{ \quad - Limited WEP} &  76.0& \multicolumn{1}{l|}{42.4}      &  71.0& \multicolumn{1}{l|}{40.1}      & 58.0& \multicolumn{1}{l|}{66.1}      &   68.3&   49.5\\ 
\multicolumn{1}{l|}{ \quad - No State Reward} &     78.0& \multicolumn{1}{l|}{39.4}      &    73.0& \multicolumn{1}{l|}{40.2}      &    \textbf{63.0}& \multicolumn{1}{l|}{53.7}      &        71.3&     44.5\\ 
\multicolumn{1}{l|}{ \quad - Frozen belief model}            &    66.5& \multicolumn{1}{l|}{43.5}      &     59.5& \multicolumn{1}{l|}{46.2}      &    28.0& \multicolumn{1}{l|}{80.3}      &        51.3&     56.7\\ 
\multicolumn{1}{l|}{ \quad - Summary-Belief}               &   60.4& \multicolumn{1}{l|}{24.0}      &     54.3& \multicolumn{1}{l|}{34.1}      &   3.5& \multicolumn{1}{l|}{18.3}      &     39.4&     25.5\\ \midrule
\multicolumn{9}{c}{\textbf{Qwen3-4B-Instruct}}                                                                                                                                                       \\ \midrule
\rowcolor{AgentBRACERow}\multicolumn{1}{l|}{\shortname}          &    \textbf{88.0}& \multicolumn{1}{l|}{30.5}      &    \textbf{81.0}& \multicolumn{1}{l|}{30.0}      &     \textbf{69.0}& \multicolumn{1}{l|}{44.6}      &       \textbf{79.3}&  35.1\\
\multicolumn{1}{l|}{ \quad - Limited WEP} &   79.0  & \multicolumn{1}{l|}{35.6}      &  69.0  & \multicolumn{1}{l|}{40.7}      &  48.0  & \multicolumn{1}{l|}{73.9}      &   65.3     &  50.1   \\ 
\multicolumn{1}{l|}{ \quad - No State Reward} &   59.5& \multicolumn{1}{l|}{44.4}      &  64.5& \multicolumn{1}{l|}{45.1}      &  58.5& \multicolumn{1}{l|}{57.6}      &   60.8&      45.0\\ 
\multicolumn{1}{l|}{ \quad - Frozen belief model}            &  77.5& \multicolumn{1}{l|}{34.7}      &  57.0& \multicolumn{1}{l|}{50.1}      & 35.5& \multicolumn{1}{l|}{74.1}      &  56.7&     53.3\\ 
\multicolumn{1}{l|}{ \quad - Summary-Belief}               & 61.5& \multicolumn{1}{l|}{47.1}      &  36.0& \multicolumn{1}{l|}{66.4}      &  38.5& \multicolumn{1}{l|}{67.4}      &        45.3&       60.3\\ \bottomrule
\end{tabular}
\vspace{-1em}
\label{tab:ablation_results}
\end{table}

\begin{wrapfigure}{r}{0.55\textwidth}
    \centering
    \vspace{-1em}
    \includegraphics[width=0.54\textwidth]
    {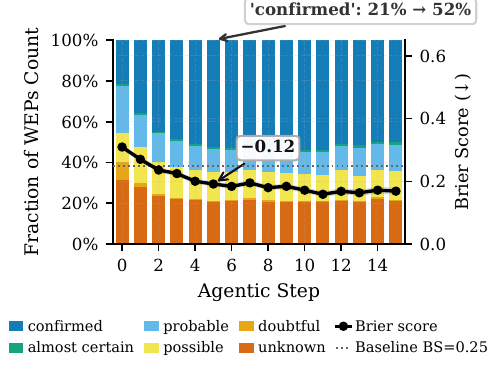}
    \vspace{-0.5em}
    \caption{Brier score drops from 0.40 $\rightarrow$ 0.28 while \textit{confirmed} claims grow 21\% to 52\%, confirming progressive calibration as evidence accumulates. WEP label distribution (bars, left axis) and mean Brier Score (line, right axis) across agent steps for \qwenthree (\shortname) on Quest dataset.
    }
    \label{fig:brier_score}
\vspace{-0.1em}
\end{wrapfigure}

\textbf{Belief uncertainty becomes better calibrated over the course of an episode.} \cref{fig:brier_score} analyzes the uncertainty representations learned by the belief model. To compute the Brier scores, we map each WEP label to its nominal probability following the ordinal scale (e.g., \textit{confirmed} $\approx$ 1.0, \textit{probable} $\approx$ 0.75, \textit{possible} $\approx$ 0.50, \textit{doubtful} $\approx$ 0.25, \textit{unknown} $\approx$ 0.0), and compute the Brier score~\citep{glenn1950verification}  against binary ground truth from the LLM judge (Qwen3-30B-A3B-Instruct-2507); 0 indicates perfect calibration and 0.25 corresponds to random chance brier score.
Brier score decreases steadily from 0.4 at step 0 to below 0.28 by step 14, a reduction of approximately 0.12 points, and remains well below the random baseline of 0.25 throughout, while the fraction of \textit{confirmed} claims grows from 21\% to 52\% -- confirming that the belief model progressively sharpens its uncertainty estimate as evidence accumulates. 
Additional calibration analysis (Appendix~\ref{app:belief_underconfident}) reveals that the belief model begins underconfident but becomes increasingly calibrated over the course of training, 
with high-confidence labels (\textit{almost certain}, \textit{confirmed}) remaining well-calibrated throughout -- a safer failure mode than overconfidence~\citep{stengel2024lacie}. Appendix~\ref{app:example} provides qualitative examples of uncertainty labels being correctly assigned and progressively resolved as evidence accumulates.

\textbf{Joint training of belief model and policy drives performance gains.}
\cref{tab:ablation_results} presents the results of various ablations of \shortname on \qwentwo and \qwenthree. When the belief model is frozen and restricted to summarizing past observations (Summary-Belief), the average performance drops by 33.4\% and 34.0\% for \qwentwo and \qwenthree, respectively. The drop is most severe on Cooking (58.5\% $\rightarrow$ 3.5\% on \qwentwo; 69.0\% $\rightarrow$ 38.5\% on \qwenthree), because Cooking requires fine-grained uncertainty tracking over sequential sub-goals and collapsing the belief distribution is particularly damaging when the agent must reason over multiple possible world states simultaneously. Similarly, freezing the belief model and training only the policy (Frozen belief model) shows an average accuracy drop by 21.5\% absolute points on \qwentwo and 22.6\% points on \qwenthree, demonstrating the importance of optimizing the belief and policy models rather than treating belief as a fixed module.

\textbf{Belief quality depends on reward design, label granularity, and task-specific supervision.} 
As seen in ~\cref{tab:ablation_results}, removing the state-grounding rewards (state correctness, state tracking, diversity, and format) degrades average accuracy by 1.5\% on \qwentwo and 18.5\% on \qwenthree, confirming that explicit grounding signals are critical for downstream performance. Collapsing the 7-level WEP scale to a binary certain/unknown drops average accuracy from 79.3\% $\rightarrow$ 65.3\% on \qwenthree and 72.8\% $\rightarrow$ 68.3\% on \qwentwo,
demonstrating that uncertainty granularity directly impacts policy performance. Additionally we also ablated task-specific belief supervision (Appendix~\ref{app:belief_without_task}), i.e, both the SFT warm-start and domain-structured prompting, which causes a moderate drop from 79.3\% $\rightarrow$ 69.3\% on \qwenthree, yet without belief supervision, \shortname still outperforms most trained baselines, confirming that joint RL alone provides sufficient signal for structured world-state tracking.

\section{Related Work}
Belief state estimation has a long history in sequential decision making under partial observability. Classical POMDP solvers~\citep{kaelbling1998planning, smallwood1973optimal} maintain exact belief distributions over finite, discrete state spaces with known transition and observation models. To scale to continuous state spaces, Monte Carlo methods such as particle filters approximate the belief distribution~\citep{silver2010monte, thrun1999monte}. When transition and observation models are unknown, deep RL approaches learn belief representations as distributions over latent states via variational inference~\citep{hafner2020mastering, gregor2019shaping}. However, these approaches assume a pre-specified, continuous state space and learn implicit latent representations, assumptions that break down in text-based environments where the state space is open-ended and interpretability is desirable.

Increasingly, LLMs form the backbone of interactive agents for long-horizon, partially observable tasks such as software and web navigation.
ReAct~\citep{yao2022react} interleaves chain-of-thought reasoning with actions, conditioning on a growing interaction history. 
Several memory-based approaches~\citep{yu2025memagent, kang2025acon, zhang2025memory, xu2025mem} address the resulting context burden: MEM1~\citep{zhou2025mem1} selectively compresses past interactions through summarization. Although these methods reduce the burden of long contexts, they still treat history as a sufficient representation of the agent's state, without modeling uncertainty over the environment. 

A complementary line of work directly addresses belief representation in LLM agents. One approach uses internal latent states as an implicit proxy for belief~\citep{kamel2025emergent}, sacrificing interpretability. At the other extreme, StateAct \citep{rozanov2025stateact}, ABBEL \citep{lidayan2025abbel}, and PABU~\citep{jiang2026pabu} represent agent state as structured natural language summaries, but collapse the belief distribution to a single maximum-likelihood estimate, discarding uncertainty inherent to belief states. \shortname addresses it by externalizing belief in structured text while preserving uncertainty via per-claim WEP annotations and jointly optimizing the belief state model and the policy model via reinforcement learning.

\section{Conclusion}
We introduce Agent-BRACE, a training method that jointly trains a belief state model and a policy model via reinforcement learning. The belief state model produces a structured approximation of the belief distribution as atomic natural language claims annotated with a Words of Estimative Probability label on an ordered scale. Conditioning the policy on this compact belief rather than the raw interaction history simultaneously addresses two challenges that arise when LLM agents operate in POMDP-style environments: representing uncertainty over an open-ended state space, and bounding the context required for action selection. \shortname attains the highest average performance on three TextWorld environments on both \qwentwo and \qwenthree.
Despite training only on Quest, the method transfers to the held-out Treasure and Cooking tasks, suggesting the structured belief representation captures task-relevant attributes without per-task engineering. Ablations confirm that joint optimization, the graded WEP scale, and the state tracking/correctness rewards each contribute meaningfully. Further analysis shows the learned belief is well calibrated and sharpens as evidence accumulates, with Brier score dropping as the steps progress.

\section*{Acknowledgments}
This work was supported by Microsoft Agentic AI Research and Innovation (AARI) grant program, NDSEG PhD Fellowship, NSF-AI Engage Institute DRL-2112635, NSF-CAREER Award 1846185, and an Apple PhD Fellowship. The views contained in this article are those of the authors and not of the funding agency.

\bibliography{ref_conference}
\bibliographystyle{plainnat}

\appendix

\newpage

\section{Belief State Structure} 
\label{app:belief_structure}
In \textit{TextWorld} environment, the belief state tracks five critical dimensions: (i) the agent's current location, (ii) topological room connections, (iii) states of observed objects, (iv) inventory contents,  and (v) progress relative to specific sub-goals. To ensure a clean separation of concerns between the belief model and the policy, the belief state is strictly prohibited from generating plans, intentions, or hypothetical future actions. This ensures that $b_k$ serves exclusively as an approximation of the current environment states, leaving the decision making to the policy model. ~\cref{exp:belief_state_prompt} shows the full prompt used for generating the belief state.

\section{Belief Model SFT on Teacher Trajectories}
\label{app: SFT}
We first perform SFT on the belief model $f_\phi$ using a teacher dataset $\mathcal{D}_{\text{teacher}}$. These trajectories are generated by GPT 5.4 mini using Prompt~\ref{exp:belief_state_prompt}, which is prompted to perform state tracking. For each transition $(G, b_{t-1}, o_t) \to b_k$, we optimize the standard cross entropy loss:
\begin{equation}
    \mathcal{L}_{\text{SFT}}(\phi) = - \mathbb{E}_{(g, b_{k-1}, o_k, b_k) \sim \mathcal{D}_{\text{teacher}}} \left[ \sum_{t} \log f_\phi (b_{k,t} \mid g, b_{k-1}, o_k, b_{k,<t}) \right]
\end{equation}
This stage is necessary to teach the belief model to utilize the Likert certainty scale and adhere to the structured representation.  

To verify that the SFT phase in belief state model training only bootstraps structural formatting rather than transferring task knowledge, ~\cref{fig:belief_model_curve} tracks individual reward components across belief state model training (Appendix~\ref{app:belief_model_curve}). Format compliance remains consistently high (>0.96) from the first training step,  while state correctness, diversity, and task rewards all begin near their lower bounds and rise steadily, confirming that belief quality is learned through RL, not inherited from the SFT teacher.

\section{Belief State without Task-Specific Supervision}
\label{app:belief_without_task}
To decouple the contribution of task-specific belief supervision, we run an additional ablation: \shortname without belief supervision. First, the belief state model is initialized directly from the base model (\qwenthree) rather than from an SFT checkpoint trained on belief state annotations from GPT 5.4 mini, eliminating the supervised warm-start. Second, the belief state prompt template is made fully domain-agnostic: all game-specific entity names like room and object identifiers in the in-context examples are either removed or replaced with abstract placeholders, removing the domain-specific belief prompt (ref. ~\cref{exp:belief_state_prompt_generic} for full prompt). Under this configuration, the belief state model must bootstrap a meaningful world-state representation entirely through PPO training, guided only by the task reward signal and belief state model rewards.

As shown in ~\cref{tab:belief_state_generic}, removing belief supervision leads to consistent degradation across all three tasks -- average accuracy drops from 79.3\% to 69.3\%. Comparing against the full \qwenthree baseline suite from ~\cref{tab:main_result}, the ablation (- Belief supervision) at 69.3\% average accuracy still outperforms PABU (62.7\%), ReAct RL (54.2\%), MEM1 (45.0\%), ReAct (47.8\%), and the Base Model (65.3\%), falling only a little behind Direct-Action RL (74.0\%) among trained baselines. Critically, the degradation here is moderate rather than catastrophic: the system is still able to retain substantial task-solving capability even without any belief supervision. This confirms that the SFT warm-start and domain-structured prompting are important catalysts for belief state model learning, but the joint RL training alone can provide a sufficiently strong training signal to teach structured world-state tracking from a general-purpose language prior. This result is encouraging for real-world applicability, where it is not always feasible to define task-relevant states.

\begin{table}[!h]
\caption{Ablation of belief state supervision on \qwenthree. \shortname vs. a variant with no SFT initialization and no domain-specific belief prompt structure (- Belief Supervision).}
\centering
\small
\begin{tabular}{lllllllll}
\toprule
\multicolumn{1}{l|}{Method}              & \multicolumn{2}{c|}{Quest}            & \multicolumn{2}{c|}{Treasure}         & \multicolumn{2}{c|}{Cooking}          & \multicolumn{2}{c}{Average} \\ \midrule
\multicolumn{1}{l|}{}                    & Acc.$\uparrow$ & \multicolumn{1}{l|}{Steps$\downarrow$} & Acc.$\uparrow$ & \multicolumn{1}{l|}{Steps$\downarrow$} & Acc.$\uparrow$ & \multicolumn{1}{l|}{Steps$\downarrow$} & Acc.$\uparrow$       & Steps$\downarrow$      \\ \midrule
\rowcolor{AgentBRACERow}\multicolumn{1}{l|}{\shortname}          &    \textbf{88.0}& \multicolumn{1}{l|}{30.5}      &    \textbf{81.0}& \multicolumn{1}{l|}{30.0}      &     \textbf{69.0}& \multicolumn{1}{l|}{44.6}      &       \textbf{79.3}&  35.1\\
\multicolumn{1}{l|}{ \quad - Belief Supervision}               & 79.0 & \multicolumn{1}{l|}{34.9}      & 74.0 & \multicolumn{1}{l|}{38.1}      & 55.0 & \multicolumn{1}{l|}{62.4}      &  69.3      & 45.1      \\ \bottomrule
\end{tabular}
\label{tab:belief_state_generic}
\end{table}

\section{Reward Design}
\label{app:reward_design}

To train \shortname, we define two distinct reward signals: 
one for the belief state model ($R^{\text{belief}}$) and one for the policy model($R^{\text{env}}$) -- which are optimized jointly so that the belief representation is shaped by the policy's decision-making needs.

\paragraph{Belief State Model Rewards ($R_t^{\text{belief}}$):} As shown in ~\cref{fig:method}, the belief state model is trained using a composite reward consisting of five signals. To show the importance of these rewards, we ablate belief state relevant rewards in~\cref{sec:ablation}. 

\begin{itemize}[leftmargin=1em]
    \item \textbf{State Tracking Reward} ($r^{\text{st}}_t$): This reward assesses the logical consistency of the belief update $b_{t-1} \to b_t$ given observation $o_t$. An LLM judge (ref. ~\cref{exp:state_tracking_reward_prompt}) assesses whether new information in $o_k$ is incorporated in the updated belief. This reward counts, $N_{\text{new}}$: new facts in $o_t$ correctly added; $N_{\text{missing}}$: new facts absent or wrong in $b_t$; $N_{\text{stale}}$: prior beliefs contradicted by $o_t$ but left unchanged; $N_{\text{total}}$: total claims in $b_t$. The reward is the product of coverage of new information and freshness of retained beliefs: \(r^{\text{st}}_t = \frac{N{\text{new}}}{N_{\text{new}}+N_{\text{missing}}}\times\Bigl(1-\frac{N_{\text{stale}}}{N_{\text{total}}}\Bigr).\)

    \item \textbf{State Correctness Reward} ($r^{\text{sc}}_t$): This reward ensures that the belief state claim, along with its uncertainty score, remains grounded to the environment states. An LLM judge parses $b_t$ into tuples $(subject, predicate, certainty)$ (ref. ~\cref{exp:claim_extraction_prompt}), then verifies each tuple against $s_t$, classifying it as fully correct, partially correct (the underlying fact is true, but the certainty label is miscalibrated) and incorrect (ref. ~\cref{exp:claim_verification_prompt}). Therefore final $r^{\text{sc}}_t = \frac{N{\text{correct}} + 0.5 \times N_{\text{partial}}}{N_{\text{ver}}}$, where $N_{\text{ver}}=N_{\text{correct}}+N_{\text{partial}}+N_{\text{incorrect}}$. 
    \item \textbf{Format Reward} ($r^{\text{format}}_t$): This reward assess adherence to the structured representation. To mitigate reward hacking and degradation of the structure, we apply a multiplicative gating signal that enforces these constraints, assigning zero reward to structurally invalid outputs. 
    \item \textbf{Diversity Reward} ($r^{\text{div}}_t$): Each uncertainty score in the belief state is matched against an ordered keyword list (e.g., \textit{confirmed} $\approx$ 1.0, \textit{probable} $\approx$ 0.75, \textit{possible} $\approx$ 0.50, \textit{doubtful} $\approx$ 0.25, \textit{unknown} $\approx$ 0.0) and mapped to one of 7 canonical levels. Shannon entropy $H = -\sum p_i \log p_i$ is computed over the resulting label distribution and normalized to $[0, 1]$.
    The reward is maximized when each individual claim's uncertainty is spread evenly across the WEP vocabulary and minimized when they collapse onto a single label.
    \item \textbf{Discounted Success Reward} ($r^{\text{success}}_t$)  A time step-discounted success reward ($\gamma^t \times \mathbf{1}[\text{success}]$) is assigned to each belief state based on the agent's final task outcome.
\end{itemize}
The final reward assigned to a belief state is \(R^{\text{belief}}_t = r^{\text{format}}_t \times \frac{1}{4}\bigl(r^{\text{st}}_t + r^{\text{sc}}_t + r^{\text{div}}_t + r^{\text{success}}_t\bigr)\). The format reward acts as a multiplicative gate, zeroing out all other rewards for structurally invalid outputs and preventing reward hacking. 

\paragraph{Policy Model Reward ($R^{\text{env}}$):} The policy model $\pi_\theta$ is optimized to maximize the policy reward, which is derived from the external environment. This signal (+1 for task success, 0 for task failure) provides the primary reinforcement for action selection, ensuring that the policy learns to take the next best action.

\section{Dataset Details}
\label{app:dataset}

All tasks are built on the \textit{TextWorld} environment, a procedural text-game generator that produces fully observable game graphs alongside natural-language descriptions. We construct three benchmark suites with controlled difficulty curricula: Quest, Treasure, and Cooking. All three suites share the same split sizes: 1,000 training games, 100 validation games, and 200 test games, generated with non-overlapping base seeds (10000 / 20000 / 30000, respectively). Training games span easier difficulty levels, while validation and test games sample progressively harder configurations to measure out-of-distribution generalization. We also extend our dataset to the ALFWorld environment.

\subsection{Quest}
\paragraph{Task.} The agent must navigate a multi-room environment, locate and manipulate objects (keys, containers, doors), and collect a designated target object. Winning requires executing the full quest sequence in the correct order.

\paragraph{Generation.}
Games are generated with \texttt{tw-make custom}, parametrized by a \texttt{rooms:objects:quest\-length} triplet that directly controls world complexity. Training games cycle across four configurations (Table~\ref{tab:basic_configs}). Validation and test games cycle across five harder configurations ranging from 6 rooms\,/\,6 objects\,/\,8-step quests up to 8 rooms\,/\,12 objects\,/\,13-step quests.

\begin{table}[h]
\centering
\caption{Basic (Quest) training configurations.}
\begin{tabular}{cccc}
\toprule
\textbf{Configuration} & \textbf{Rooms} & \textbf{Objects} & \textbf{Quest Steps} \\
\midrule
2:3:3 & 2 & 3  & 3  \\
2:4:4 & 2 & 4  & 4  \\
4:4:4 & 4 & 4  & 4  \\
4:6:6 & 4 & 6  & 6  \\
\bottomrule
\end{tabular}
\label{tab:basic_configs}
\end{table}

\subsection{Treasure}
\label{app:dataset:treasure}

\paragraph{Task.}
The agent is placed inside a procedurally generated maze and must locate a named treasure object (\emph{e.g.}, a latchkey) hidden in a random room. The objective is given in natural language at the start of each episode.

\paragraph{Generation.}
Games are generated with \texttt{tw-make tw-treasure\_hunter --level} $L$, where the level integer ($1$--$30$) jointly governs world size, container nesting depth, and the number of distractor objects. Training games cycle across levels $\{1, 2, 4, 6, 8\}$ (easy band); Validation and test games cycle across levels $\{14, 16, 18, 20, 22, 25, 28, 30\}$.

\subsection{Cooking}
\label{app:dataset:cooking}

\paragraph{Task.}
The agent must find a recipe posted in a kitchen cookbook and execute it: navigate to relevant rooms, gather the required ingredients, apply the correct preparation steps (opening containers, cutting, cooking), and finally prepare and eat the meal.

\paragraph{Generation.}
Games are generated with \texttt{tw-make tw-cooking}, parametrized as \texttt{recipe:take:go:flags}, where \texttt{recipe} is the number of required ingredients, \texttt{take} is the number of objects to pick up, \texttt{go} is the number of rooms, and \texttt{flags} encode which mechanics are active: \texttt{o} (openable containers), \texttt{c} (cooking appliance), \texttt{t} (cutting board), \texttt{d} (limited inventory\,/\, drop required). Training configurations cycle across four settings (Table~\ref{tab:cooking_configs}). Validation and test configurations cycle across five harder settings (4--5 ingredients, 9--12 rooms).

\begin{table}[h]
\centering
\caption{Cooking training configurations.}
\begin{tabular}{lcccc}
\toprule
\textbf{Configuration} & \textbf{Rooms} & \textbf{Ingredients} & \textbf{Mechanics} \\
\midrule
3:3:9:oc   & 9  & 3 & open, cook              \\
4:4:6:oct  & 6  & 4 & open, cook, cut         \\
4:4:6:octd & 6  & 4 & open, cook, cut, drop   \\
4:4:9:oct  & 9  & 4 & open, cook, cut         \\
\bottomrule
\end{tabular}
\label{tab:cooking_configs}
\end{table}

\subsection{ALFWorld}
The training and testing datasets for ALFWorld are directly taken from ~\cite{wang2025practitionersguidemultiturnagentic}.

\section{Implementation Details and Hyperparameters}
\label{app:hyperparameters}
Our codebase is built on top of ~\cite{wang2025practitionersguidemultiturnagentic} and all training is run on a single node with 4 NVIDIA GPUs (A100).

\subsection{Models Used}
To train and evaluate \shortname, we adopt instruction-tuned models, \qwentwo\footnote{https://huggingface.co/Qwen/Qwen2.5-3B-Instruct} and \qwenthree\footnote{https://huggingface.co/Qwen/Qwen3-4B-Instruct-2507} as our base models. Additionally, we use Qwen3-30B-A3B-Instruct-2507\footnote{https://huggingface.co/Qwen/Qwen3-30B-A3B-Instruct-2507} as the judge to evaluate state tracking and correctness reward.

\subsection{Belief State Model: Supervised Pre-training (SFT)}
\label{app:belief-sft}

Before joint RL training, the belief-state LM is fine-tuned with supervised learning on belief-state trajectories generated from teacher demonstrations. Hyperparameters are summarised in Table~\ref{tab:sft-hparams}.

\begin{table}[h]
\centering

\caption{Belief-state SFT hyperparameters.}
\begin{tabular}{lc}
\toprule
\textbf{Hyperparameter} & \textbf{Value} \\
\midrule
Base model & Qwen3-4B-Instruct \\
Max sequence length & 4096 \\
Training epochs & 5 \\
Per-device batch size & 4 \\
Gradient accumulation steps & 8 \\
Effective batch size & 32 \\
Learning rate & $2\times10^{-5}$ \\
Precision & BF16 \\
LoRA rank ($r$) & 64 \\
LoRA $\alpha$ & 128 \\
Save steps & 100 \\
\bottomrule
\end{tabular}
\label{tab:sft-hparams}
\end{table}

\subsection{Policy PPO Training}
\label{app:ppo-hparams}

The policy is trained with Proximal Policy Optimization (PPO) using a Generalized Advantage Estimation (GAE) critic. Table~\ref{tab:ppo-hparams} lists the PPO hyperparameters used.

\begin{table}[h]
\centering

\caption{Policy PPO hyperparameters (shared across environments).}
\begin{tabular}{lc}
\toprule
\textbf{Hyperparameter} & \textbf{Value} \\
\midrule
\multicolumn{2}{l}{\textit{Algorithm}} \\
Advantage estimator & GAE \\
Discount factor $\gamma$ & 1.0 \\
KL penalty coefficient & 0.01 \\
Clip ratio $\varepsilon$ & 0.2 \\
\midrule
\multicolumn{2}{l}{\textit{Optimization}} \\
Actor learning rate & $1\times10^{-6}$ \\
Critic learning rate & $1\times10^{-5}$ \\
Train batch size & 16 \\
PPO mini-batch size & 16 \\
\midrule
\multicolumn{2}{l}{\textit{Rollout}} \\
Rollout temperature & 0.7 \\
Validation temperature & 0.4 \\
Rollout samples per prompt ($n$) & 1 \\
Policy tensor-parallel size & 2 \\
Maximum episode steps (training) & 15 \\
Maximum episode steps (evaluation) & 100 \\
Reward shaping & single (terminal) \\
\bottomrule
\end{tabular}
\label{tab:ppo-hparams}
\end{table}

\subsection{Joint Belief-State PPO Training}
\label{app:belief-rl-hparams}

During RL, the belief-state LM is updated jointly with the policy after every rollout batch. Belief-model training hyperparameters are given in Table~\ref{tab:belief-rl-hparams}.

\begin{table}[h]
\centering

\caption{Joint belief-state RL hyperparameters.}
\begin{tabular}{lc}
\toprule
\textbf{Hyperparameter} & \textbf{Value} \\
\midrule
\multicolumn{2}{l}{\textit{Optimization}} \\
\midrule
Belief LM learning rate & $1\times10^{-5}$ \\
Gradient update steps per batch & 1 \\
PPO mini-batch size & 4 \\
PPO clip ratio $\varepsilon$ & 0.2 \\
Dual-clip upper bound $c$ & 3.0 \\
Policy loss mode & vanilla PPO \\
Value function & disabled \\
\midrule
\multicolumn{2}{l}{\textit{Inference (vLLM subprocess)}} \\
\midrule
Max generation tokens & 1{,}024 \\
vLLM max model length & 4{,}096 \\
vLLM max batched tokens & 4{,}096 \\
\bottomrule
\end{tabular}
\label{tab:belief-rl-hparams}
\end{table}

\section{Belief States are underconfident but improve over training}
\label{app:belief_underconfident}
\begin{figure}[!h]
    \centering
    \includegraphics[width=0.8\textwidth]{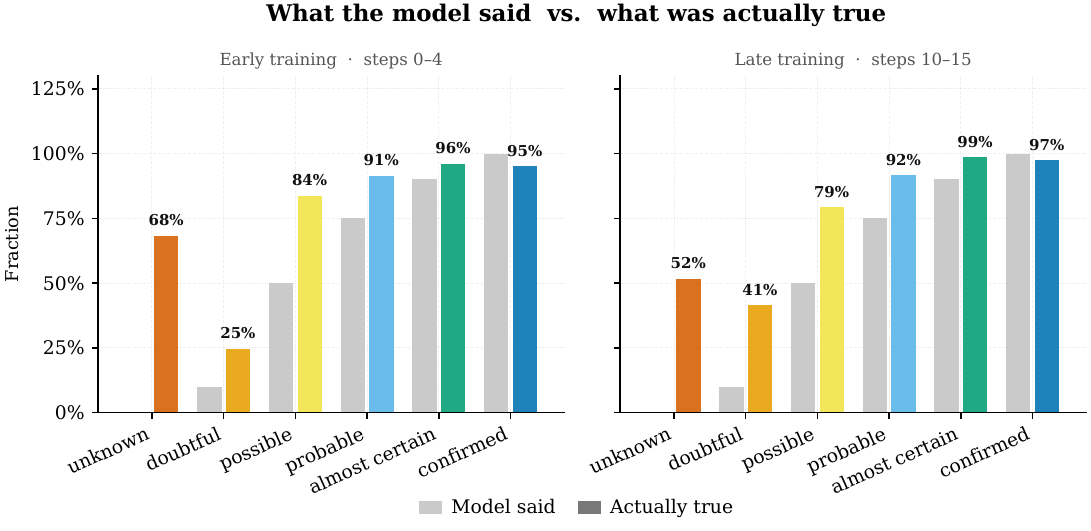}
    \caption{Calibration of WEP labels at early (0-4) and late (10-15) steps. For each WEP label emitted by the belief model, the colored bar shows the fraction of those claims independently verified as true, while the grey bar shows the nominal probability that the label represents on the WEP scale. The belief model is systematically underconfident; it assigns conservative labels to beliefs that are more often true than the label implies.}
    \label{fig:underconfident}
\end{figure}

\cref{fig:underconfident} analyzes the calibration of WEP labels at early (steps 0-4) and late (steps 10-15) training stages. For each WEP label emitted by the belief state model, we measure the empirical truth rate -- the fraction of claims carrying that label that are independently verified as true by the LLM judge (Qwen3-30B-A3B-Instruct-2507) and compare it against the \textit{nominal probability} that the label represents on the WEP scale (e.g., \textit{confirmed} $\approx$ 1.0, \textit{probable} $\approx$ 0.75, \textit{possible} $\approx$ 0.50, \textit{doubtful} $\approx$ 0.25, \textit{unknown} $\approx$ 0.0; shown in gray). 
A perfectly calibrated belief model would have these match exactly; a model whose empirical truth rate consistently exceeds the nominal probability is underconfident; it assigns lower-confidence labels to beliefs that are in fact more often true than those labels suggest (e.g., labeling something \textit{possible} when it is actually true 84\% of the time). ~\cref{fig:underconfident} shows that this is precisely what occurs: at early training, claims labeled \textit{unknown} are verified true 68\% of the time, far above the nominal probability of approximately 0, shrinking to 54\% by late training. This indicates the model increasingly reserves \textit{unknown} for genuinely uncertain claims rather than as a default label. At the high-confidence end, \textit{almost certain} and \textit{confirmed} remain well calibrated throughout (
$\geq$91\% and $\geq$95\% respectively). In sequential decision making under partial observability, underconfidence is a safer failure mode than overconfidence~\citep{stengel2024lacie}: an agent that hedges will continue to explore and gather evidence, whereas an overconfident agent risks committing to an incorrect world model and acting on it irreversibly. 

\section{Statistical Reliability of Main Results}
\label{app:error}
To assess the statistical reliability of \shortname, we report mean accuracy and standard deviation across three independent runs for all methods trained on Quest and evaluated on all three TextWorld environments. ~\cref{tab:error_table} reports the result for \qwenthree across three independent runs. 

\begin{table}[!h]
\centering
\small
\caption{
Performance comparison of \shortname against baselines across three TextWorld
environments (Quest, Treasure, Cooking) on \qwenthree with standard deviation across three independent runs.}
\begin{tabular}{lllllll}
\toprule
\multicolumn{1}{l|}{Method} & \multicolumn{2}{c|}{Quest} & \multicolumn{2}{c|}{Treasure} & \multicolumn{2}{c}{Cooking} \\ \midrule
\multicolumn{1}{l|}{} & Acc.$\uparrow$ & \multicolumn{1}{l|}{Steps$\downarrow$} & Acc.$\uparrow$ & \multicolumn{1}{l|}{Steps$\downarrow$} & Acc.$\uparrow$ & \multicolumn{1}{l}{Steps$\downarrow$} \\ \midrule
\rowcolor{InferenceRow}\multicolumn{1}{l|}{Base Model}         & 61.5 $\pm$ 0.0 & \multicolumn{1}{l|}{32.2 $\pm$ 0.1} & 66.0 $\pm$ 0.9 & \multicolumn{1}{l|}{29.9 $\pm$ 0.3} & 69 $\pm$ 1.4 & \multicolumn{1}{l}{33.9 $\pm$ 0.2} \\
\rowcolor{InferenceRow}\multicolumn{1}{l|}{ReAct}              & 61.0 $\pm$ 0.6 & \multicolumn{1}{l|}{12.8 $\pm$ 0.2} & 69.5 $\pm$ 0.9 & \multicolumn{1}{l|}{10.0 $\pm$ 0.2} & 13.2 $\pm$ 0.3 & \multicolumn{1}{l}{24.4 $\pm$ 0.0}  \\
\multicolumn{1}{l|}{Direct-Action (RL)}                        & 74 $\pm$ 0.8 & \multicolumn{1}{l|}{29.8 $\pm$ 0.7} & 72.5 $\pm$ 0.3 & \multicolumn{1}{l|}{28.1 $\pm$ 0.1} & 76 $\pm$ 0.6 & \multicolumn{1}{l}{31.5 $\pm$ 0.3} \\
\multicolumn{1}{l|}{ReAct (RL)}                                & 75.5 $\pm$ 0.6 & \multicolumn{1}{l|}{18.3 $\pm$ 0.2} & 74.0 $\pm$ 0.0 & \multicolumn{1}{l|}{16.5 $\pm$ 0.0} & 13.0 $\pm$ 0.0 & \multicolumn{1}{l}{40.6 $\pm$ 0.0}  \\
\multicolumn{1}{l|}{PABU}                                      & 83.0 $\pm$ 0.9 & \multicolumn{1}{l|}{26.4 $\pm$ 2.4} & 70.7 $\pm$ 3.0 & \multicolumn{1}{l|}{37.8 $\pm$ 1.4} & 32.5 $\pm$ 0.3 & \multicolumn{1}{l}{72.1 $\pm$ 3.1} \\
\multicolumn{1}{l|}{\shortname}        & 88 $\pm$ 0.9 & \multicolumn{1}{l|}{30.8 $\pm$ 0.5} & 81 $\pm$ 0.6 & \multicolumn{1}{l|}{29.9 $\pm$ 0.2} & 68.7 $\pm$ 0.6 & \multicolumn{1}{l}{44.7 $\pm$ 0.2} \\ \toprule
\end{tabular}
\label{tab:error_table}
\end{table}

\section{Belief State Model Training}
\label{app:belief_model_curve}
\begin{figure}[!h]
    \centering
    \includegraphics[width=\textwidth]{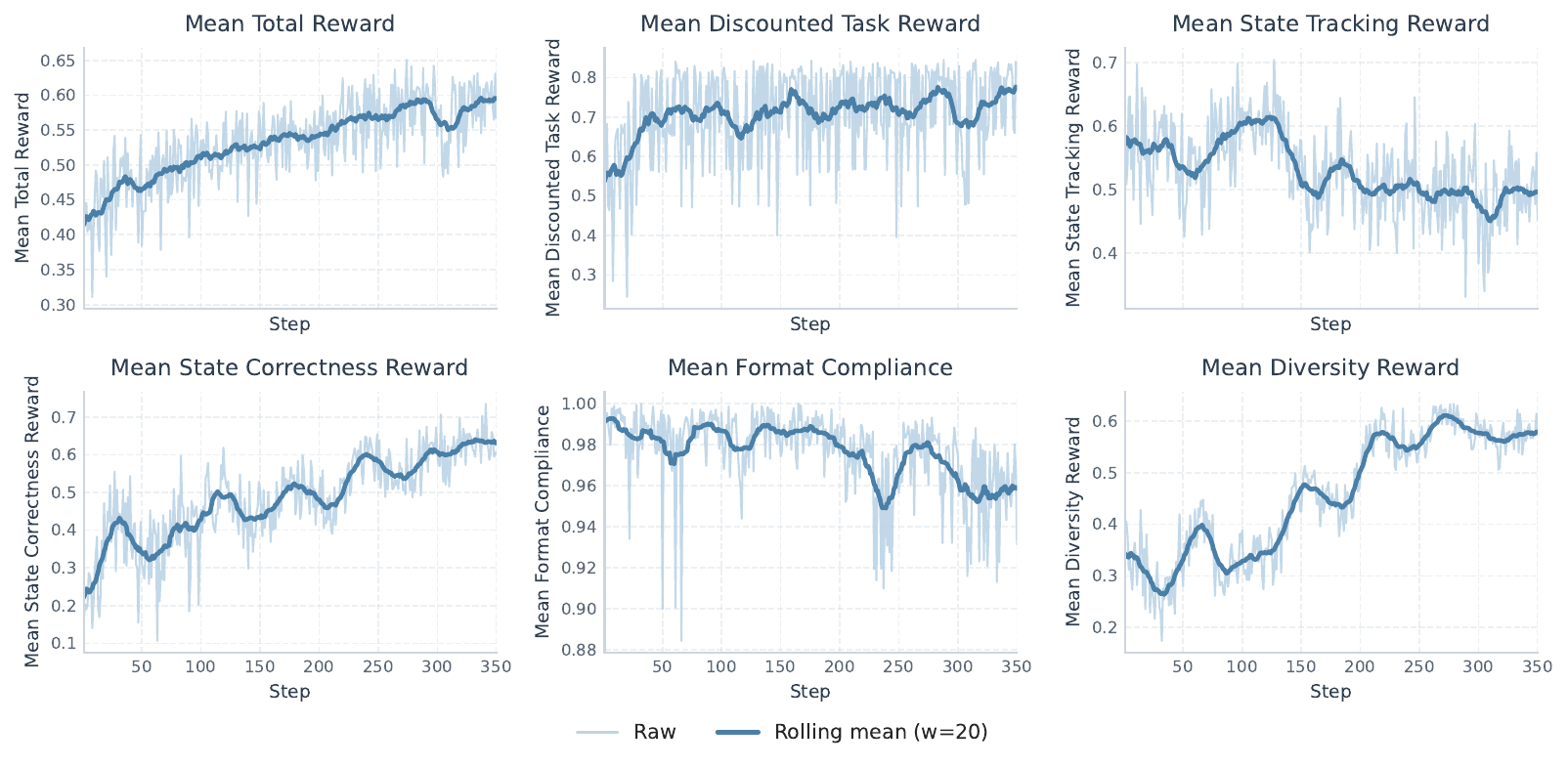}
    \caption{Reward component trajectories across PPO training steps for the belief state model (\qwenthree). }
    \label{fig:belief_model_curve}
\end{figure}

~\cref{fig:belief_model_curve} plots the five reward components along with the mean total reward across belief state model training steps. If the SFT cold start phase were distilling task knowledge from GPT 5.4 mini, we would expect state correctness and total task reward to begin high and plateau quickly, with minimal learning signal during PPO. Instead, we observe the opposite: format compliance starts near 0.98 and remains consistently high throughout training, confirming that SFT successfully bootstrapped structured adherence to the WEP-annotated belief format. In contrast, state correctness begins near 0.2 and rises steadily to 0.65, diversity reward rises from 0.3 to 0.65, and similarly discounted task reward from 0.55 to 0.78. This pattern demonstrates that factual belief quality, uncertainty diversity, and task performance are learned through RL interaction with the environment, not inherited from the teacher model. The SFT phase contributes only to the format; all substantive learning is attributed to the RL training signal.

\section{\shortname without state correctness and tracking reward}
~\cref{tab:belief_no_llm_judge} ablates the LLM judged belief state rewards by setting $r_t^{\text{st}}$ and $r_t^{\text{sc}}$ to zero while keeping all other components fixed. Removing both the rewards degrades the average accuracy by 4\% (79.3\% $\rightarrow$ 75.3\%). Notably, on average the ablation still outperforms all baselines reported in ~\cref{tab:main_result}, demonstrating that the removal of LLM judge reward does not catastrophically degrade the performance of \shortname.

\begin{table}[!h]
\caption{Ablation of LLM judged based belief state rewards. \shortname vs. a variant with no State correctness and tracking reward.}
\centering
\small
\begin{tabular}{lllllllll}
\toprule
\multicolumn{1}{l|}{Method}              & \multicolumn{2}{c|}{Quest}            & \multicolumn{2}{c|}{Treasure}         & \multicolumn{2}{c|}{Cooking}          & \multicolumn{2}{c}{Average} \\ \midrule
\multicolumn{1}{l|}{}                    & Acc.$\uparrow$ & \multicolumn{1}{l|}{Steps$\downarrow$} & Acc.$\uparrow$ & \multicolumn{1}{l|}{Steps$\downarrow$} & Acc.$\uparrow$ & \multicolumn{1}{l|}{Steps$\downarrow$} & Acc.$\uparrow$       & Steps$\downarrow$      \\ \midrule
\rowcolor{AgentBRACERow}\multicolumn{1}{l|}{\shortname}          &    \textbf{88.0}& \multicolumn{1}{l|}{30.5}      &    \textbf{81.0}& \multicolumn{1}{l|}{30.0}      &     69.0& \multicolumn{1}{l|}{44.6}      &       \textbf{79.3}&  35.1 \\ 
\multicolumn{1}{l|}{ \quad - $r_t^{\text{st}}$ and $r_t^{\text{sc}}$ reward}               & 72.0 & \multicolumn{1}{l|}{40.2}      & 72.0 & \multicolumn{1}{l|}{36.8}      & \textbf{82.}0 & \multicolumn{1}{l|}{33.6}      &  75.3      & 36.8      \\ \bottomrule
\end{tabular}
\label{tab:belief_no_llm_judge}
\end{table}

\section{Qualitative Examples}
\label{app:example}
To illustrate the properties of our jointly trained belief state, we present representative examples from our method on cooking tasks. 

\cref{ex:uncertainty_resolved} illustrates calibrated uncertainty and its correct resolution: after a single \texttt{open plain door} action, the south exit transitions from \textit{possible} to \textit{confirmed}, while genuinely unknown quantities such as the cookbook location and shelf contents remain marked as unknown rather than being hallucinated.

\cref{ex:world_state_tracking} demonstrates that the belief state tracks world-state changes without over-committing to unseen information. After opening the sliding patio door, the model correctly upgrades that exit to \textit{confirmed open}, while unvisited north and west exits remain \textit{possible} with ``destination not yet observed'' — reflecting the agent's actual epistemic boundary.

\Cref{ex:inventory_tracking} shows fine-grained inventory tracking across multiple items. The belief state simultaneously records five items with exact processing descriptors (\textit{chopped burned}, \textit{sliced burned}) while correctly deferring on recipe requirements and counter contents, both marked \textit{unknown} pending cookbook consultation.

\begin{prompt}[Belief State Generation/Update]
\label{exp:belief_state_prompt}

You are playing a text-based game. Given the goal, your previous belief state, and the current observation, produce an updated belief state capturing what you know and how confidently you know it.

\smallskip
\textbf{Goal:} \texttt{\{goal\}} \quad
\textbf{Previous belief state:} \texttt{\{previous\_belief\_state\}} \quad
\textbf{Current observation:} \texttt{\{current\_obs\}}

\smallskip
Output \textbf{ONLY} a belief state within \texttt{<belief\_state>~</belief\_state>} tags.

\smallskip
\textbf{--- STRICT RULES ---}

\noindent Your belief state \textbf{MUST NOT} contain: (i) any next action, plan, or intention; (ii) forward-looking phrases: ``I will'', ``I should'', ``my next step'', ``I plan to'', etc.; (iii) any recommendation about which command to execute. \textbf{ONLY} record what you have already observed or can directly infer from past observations.

\smallskip
\textbf{--- UPDATE RULES ---}
\begin{itemize}[nosep, leftmargin=1.5em]
    \item If the current observation \textbf{CONFIRMS} a previous bullet $\rightarrow$ upgrade it to ``confirmed''.
    \item If the current observation \textbf{CONTRADICTS} a previous bullet $\rightarrow$ replace it immediately.
    \item Never carry forward a stale bullet that conflicts with a direct observation.
\end{itemize}

\smallskip
\textbf{--- CERTAINTY SCALE ---}

\noindent Every bullet \textbf{MUST} contain exactly one of these markers, used naturally in the sentence:

\smallskip
\noindent
\begin{tabular}{@{}ll@{}}
    \textit{confirmed / certain} & directly observed this turn \\
    \textit{almost certain}      & observed previously, no contradicting evidence since \\
    \textit{probable}            & inferred from goal structure or strong pattern \\
    \textit{possible}            & no visit yet, some contextual reason to believe \\
    \textit{unlikely}            & visited nearby rooms, no supporting evidence found \\
    \textit{doubtful}            & contradicting evidence exists \\
    \textit{unknown}           & no evidence \\
\end{tabular}

\smallskip
\textbf{--- MANDATORY COVERAGE ---}

\noindent You \textbf{MUST} cover each of the following in at least one bullet: current location; each known exit and where it leads; each goal-relevant object (its location and state); your inventory; progress toward each sub-goal.

\smallskip
\textbf{--- FORMAT ---}

\noindent One bullet per distinct fact, starting with ``-~''. Every bullet contains exactly one certainty marker. No JSON, no percentages, no key-value pairs.

\smallskip
\textbf{Examples:}
\begin{itemize}[nosep, leftmargin=1.5em]
    \item It is \textit{confirmed} that I am in the kitchen.
    \item The east exit \textit{almost certainly} leads to the hallway based on prior exploration.
    \item The key is \textit{probably} still in the living room where I last saw it.
    \item It is \textit{possible} the chest in the bedroom contains the goal item, though I have not visited.
    \item The couch is \textit{ruled out} from the bedroom -- I visited and did not observe it there.
    \item It is \textit{doubtful} the garden door is unlocked given every other door here has been locked.
    \item It is \textit{confirmed} that I am carrying only the brass key.
\end{itemize}
\end{prompt}

\begin{prompt}[State-Tracking Reward]
\label{exp:state_tracking_reward_prompt}

You are evaluating a belief-state update in a text-based game.

\smallskip
\textbf{Previous belief state:} \texttt{\{prev\_belief\}} \quad
\textbf{New observation:} \texttt{\{new\_obs\}} \quad
\textbf{New belief state:} \texttt{\{new\_belief\}}

\smallskip
\textbf{Step 1 — Identify} (brief, one line each):
\begin{itemize}[nosep, leftmargin=1.5em]
    \item \textbf{New facts:} list each distinct fact the observation reveals (e.g.\ ``player moved to kitchen'', ``door is open'')
    \item \textbf{Missing:} which of those new facts are absent or wrong in the new belief state
    \item \textbf{Stale:} which prior beliefs does the observation contradict that were left unchanged
\end{itemize}

\smallskip
\textbf{Step 2 — Count} (integers):
\begin{itemize}[nosep, leftmargin=1.5em]
    \item $N_\text{new}$ = number of new facts correctly captured
    \item $N_\text{missing}$ = number of new facts missing or wrong
    \item $N_\text{stale}$ = number of stale/contradicted priors left unchanged
    \item $N_\text{total}$ = total claims in the new belief state
\end{itemize}

\smallskip
\textbf{Step 3 — Compute:}
\begin{itemize}[nosep, leftmargin=1.5em]
    \item If $N_\text{total} = 0$: $\text{score} = 0.0$
    \item Otherwise:
    \begin{align*}
        \text{coverage}  &= \frac{N_\text{new}}{\max(1,\; N_\text{new} + N_\text{missing})} \\[4pt]
        \text{staleness} &= \frac{N_\text{stale}}{N_\text{total}} \\[4pt]
        \text{score}     &= \text{coverage} \times (1 - \text{staleness}) \quad \text{clamped to } [0.00, 1.00]
    \end{align*}
\end{itemize}

\smallskip
\noindent End with exactly \texttt{<score>X.XX</score>} where \texttt{X.XX} is a decimal in $[0.00, 1.00]$.

\end{prompt}

\begin{prompt}[Claim Extraction]
\label{exp:claim_extraction_prompt}

You are analysing a belief state from a text-based game.

\smallskip
\textbf{Belief state:} \texttt{\{belief\_state\}}

\smallskip
\textbf{Task —} Extract every specific factual claim from the belief state. List each claim on its own line using this exact format:

\smallskip
\texttt{CLAIM: <subject> | <predicate> | <certainty-label>}

\smallskip
\textbf{Examples:}
\begin{itemize}[nosep, leftmargin=1.5em]
    \item \texttt{CLAIM: player location | in the kitchen | certain}
    \item \texttt{CLAIM: key | on the table in the library | probable}
    \item \texttt{CLAIM: east exit from kitchen | leads to hallway | almost certain}
    \item \texttt{CLAIM: chest | open | possible}
\end{itemize}

\smallskip
\noindent List \textbf{ALL} claims now (one per line):

\end{prompt}

\begin{prompt}[State Correctness]
\label{exp:claim_verification_prompt}

You are verifying factual claims from a belief state against the true game world state.

\smallskip
\textbf{True game world state (ground truth):} \texttt{\{raw\_state\}} \quad
\textbf{Claims to verify:} \texttt{\{claims\}}

\smallskip
\textbf{Instructions —} For each claim decide:
\begin{itemize}[nosep, leftmargin=1.5em]
    \item \textbf{Correct} — the underlying fact is true \emph{and} the certainty label is appropriate.
    \item \textbf{Incorrect} — the underlying fact is false (label does not matter).
    \item \textbf{Partially correct} — the fact is true but the certainty label is badly miscalibrated (e.g.\ marked \textit{certain} for something only \textit{probable} is supported by evidence, or \textit{unknown} for something directly observable in the true state).
    \item \textbf{Unverifiable} — the ground truth does not contain enough information to confirm or deny the claim.
\end{itemize}

\smallskip
\textbf{Scoring —} Let:
\begin{itemize}[nosep, leftmargin=1.5em]
    \item $N_\text{verifiable}  = \text{Correct} + \text{Incorrect} + \text{Partially correct}$
    \item $N_\text{correct}     = $ number of Correct verdicts
    \item $N_\text{partial}     = $ number of Partially correct verdicts
\end{itemize}
\begin{equation*}
    \text{score} = \frac{N_\text{correct} + 0.5 \times N_\text{partial}}{N_\text{verifiable}}
    \qquad (N_\text{verifiable} = 0 \Rightarrow \text{score} = 0.0)
\end{equation*}

\smallskip
\noindent Provide a brief per-claim verdict, then end with exactly \texttt{<score>X.XX</score>} where \texttt{X.XX} is a decimal in $[0.00, 1.00]$.

\end{prompt}

\begin{prompt}[Belief State Generation/Update (No Belief Supervision)]
\label{exp:belief_state_prompt_generic}
You are playing a text-based game. Given the goal, your previous belief state, and the current observation, produce an updated belief state capturing what you know and how confidently you know it.

\smallskip
\textbf{Goal:} \texttt{\{goal\}} \quad
\textbf{Previous belief state:} \texttt{\{previous\_belief\_state\}} \quad
\textbf{Current observation:} \texttt{\{current\_obs\}}

\smallskip
Output \textbf{ONLY} a belief state within \texttt{<belief\_state>~</belief\_state>} tags.

\smallskip
\textbf{--- STRICT RULES ---}

\noindent Your belief state \textbf{MUST NOT} contain: (i) any next action, plan, or intention; (ii) forward-looking phrases: ``I will'', ``I should'', ``my next step'', ``I plan to'', etc.; (iii) any recommendation about which command to execute. \textbf{ONLY} record what you have already observed or can directly infer from past observations.

\smallskip
\textbf{--- UPDATE RULES ---}
\begin{itemize}[nosep, leftmargin=1.5em]
    \item If the current observation \textbf{CONFIRMS} a previous bullet $\rightarrow$ upgrade it to ``confirmed''.
    \item If the current observation \textbf{CONTRADICTS} a previous bullet $\rightarrow$ replace it immediately.
    \item If an expected element is not observed when new information is obtained $\rightarrow$ downgrade its likelihood.
    \item Never carry forward a stale bullet that conflicts with a direct observation.
\end{itemize}

\smallskip
\textbf{--- CERTAINTY SCALE ---}

\noindent Every bullet \textbf{MUST} contain exactly one of these markers:

\smallskip
\noindent
\begin{tabular}{@{}ll@{}}
    \textit{confirmed / certain} & directly observed this turn \\
    \textit{almost certain}      & observed previously, no contradicting evidence since \\
    \textit{probable}            & inferred from prior observations or consistent patterns \\
    \textit{possible}            & no direct evidence either way \\
    \textit{unlikely}            & weak or missing supporting evidence \\
    \textit{doubtful}            & contradicting evidence exists \\
    \textit{unknown}           & no evidence \\
\end{tabular}

\smallskip
\textbf{--- FORMAT ---}

\noindent One bullet per distinct fact, starting with ``-~''. Every bullet contains exactly one certainty marker from the scale above. No JSON, no percentages, no key-value pairs.

\smallskip
\textbf{Examples:}
\begin{itemize}[nosep, leftmargin=1.5em]
    \item It is \textit{confirmed} that I am in a situation where a surface and an object are present.
    \item An element previously observed is \textit{almost certainly} still present as there is no contradicting evidence.
    \item There is a \textit{probable} relationship between two elements based on prior observations.
    \item It is \textit{possible} that additional elements exist that have not yet been observed.
    \item An expected element is \textit{ruled out} in the current context as it is not observed.
    \item It is \textit{doubtful} that a previously assumed property holds given new evidence.
    \item It is \textit{confirmed} that I am associated with an element obtained earlier.
\end{itemize}
\end{prompt}

\begin{example}[Uncertainty correctly encoded and resolved]
\label{ex:uncertainty_resolved}
Two consecutive steps as the agent opens a door in the pantry (\texttt{game cooking\_30003}).

\smallskip
\noindent\textbf{Step A} --- \textit{Observation:}
\begin{quote}\small
You are now in the pantry. A normal one. What's that over there? It looks like it's a shelf. The shelf is wooden. However, the shelf, like an empty shelf, has nothing on it. There is a closed plain door leading south. You are carrying nothing.
\end{quote}
\textit{Belief state:}
\begin{itemize}[nosep, leftmargin=1.5em]\small
    \item It is \textit{confirmed} that I am in the pantry.
    \item The south exit is \textit{possible} through the closed plain door, as indicated by the door description.
    \item The shelf is \textit{probable} in the pantry, currently empty as observed.
    \item Whether the shelf contains any items is \textit{unknown}.
    \item It is \textit{confirmed} that I am carrying nothing.
    \item The location of the cookbook is \textit{unknown}, as it has not been observed in the pantry.
    \item Whether the recipe is accessible from here is \textit{unknown}.
\end{itemize}
\textbf{Action:} \texttt{open plain door} \checkmark

\smallskip
\noindent\textbf{Step B} --- \textit{Observation:}
\begin{quote}\small
You are now in the pantry. You open plain door. You are carrying nothing.
\end{quote}
\textit{Belief state:}
\begin{itemize}[nosep, leftmargin=1.5em]\small
    \item It is \textit{confirmed} that I am in the pantry.
    \item The south exit is \textit{confirmed} by the open plain door leading south.
    \item The shelf is \textit{probable} in the pantry, currently empty as observed.
    \item Whether the shelf contains any items is \textit{unknown}.
    \item It is \textit{confirmed} that I am carrying nothing.
    \item The location of the cookbook is \textit{unknown}, as it has not been observed here.
    \item It is \textit{confirmed} that the plain door is open, as directly observed.
\end{itemize}
\textbf{Action:} \texttt{go south} \checkmark
\end{example}

\begin{example}[Belief state tracks world-state changes across actions]
\label{ex:world_state_tracking}
The agent navigates through a corridor with a sliding patio door (\texttt{game cooking\_30002}).

\smallskip
\noindent\textbf{Step A} --- \textit{Observation:}
\begin{quote}\small
You are now in the corridor. A normal one. There is a closed sliding patio door leading east. There is an exit to the north. Don't worry, there is no door. You need an exit without a door? You should try going south. There is an exit to the west. You are carrying nothing.
\end{quote}
\textit{Belief state:}
\begin{itemize}[nosep, leftmargin=1.5em]\small
    \item It is \textit{confirmed} that I am in the corridor.
    \item The east exit is \textit{possible} through a closed sliding patio door, though it is currently closed and not open.
    \item Whether the sliding patio door can be opened is \textit{unknown}, as it has not yet been interacted with.
    \item The north exit is \textit{possible}, destination not yet observed.
    \item The west exit is \textit{possible}, destination not yet observed.
    \item It is \textit{confirmed} that I am carrying nothing.
\end{itemize}
\textbf{Action:} \texttt{open sliding patio door} \checkmark

\smallskip
\noindent\textbf{Step B} --- \textit{Observation:}
\begin{quote}\small
You are now in the corridor. You open sliding patio door. You are carrying nothing.
\end{quote}
\textit{Belief state:}
\begin{itemize}[nosep, leftmargin=1.5em]\small
    \item It is \textit{confirmed} that I am in the corridor.
    \item The east exit is \textit{possible} through an open sliding patio door, since it is explicitly open.
    \item The sliding patio door is \textit{confirmed} open, as directly observed after opening it.
    \item The north and west exits are \textit{possible}, destinations not yet observed.
    \item It is \textit{confirmed} that I am carrying nothing.
\end{itemize}
\textbf{Action:} \texttt{go east} \checkmark
\end{example}

\begin{example}[Multi-item inventory tracking with fine-grained state labels]
\label{ex:inventory_tracking}
After several cooking steps, the agent holds five distinct items (\texttt{game cooking\_30003}).

\smallskip
\noindent\textit{Observation:}
\begin{quote}\small
You are now in the kitchen. You take the red apple from the counter. You are carrying: a red apple, a knife, a chopped burned red hot pepper, a sliced burned yellow potato and a cookbook.
\end{quote}
\textit{Belief state:}
\begin{itemize}[nosep, leftmargin=1.5em]\small
    \item It is \textit{confirmed} that I am in the kitchen.
    \item The north open plain door is \textit{confirmed}.
    \item The red apple is \textit{confirmed} in my inventory, taken from the counter.
    \item The chopped burned red hot pepper is \textit{confirmed} in my inventory.
    \item The sliced burned yellow potato is \textit{confirmed} in my inventory.
    \item The cookbook is \textit{confirmed} in my inventory.
    \item The knife is \textit{confirmed} in my inventory.
    \item Whether any items remain on the counter is \textit{unknown}.
    \item The required preparation state of the red apple is \textit{unknown}, as the cookbook has not yet been read.
\end{itemize}
\textbf{Action:} \texttt{slice red apple with knife} \checkmark
\end{example}

\end{document}